\documentclass[letterpaper, 10 pt, conference]{ieeeconf}  

\IEEEoverridecommandlockouts                              
\usepackage{float} 
\usepackage{mathtools}
\usepackage{amssymb}
\usepackage{bbm}
\usepackage{graphicx}
\usepackage{multicol}
\usepackage{multirow}
\usepackage[export]{adjustbox}
\usepackage[bookmarks=true]{hyperref}
\usepackage[raggedright]{sidecap}
\usepackage[font=small,skip=5pt]{caption}
\usepackage{subcaption}
\usepackage{algorithmic}
\let\labelindent\relax
\usepackage{enumitem,kantlipsum}
\usepackage[tight-spacing=true]{siunitx}
\usepackage{tikz,tkz-graph}
\usepackage{tablefootnote}
\usepackage{booktabs}
\usepackage{dsfont}
\usepackage{gensymb}
\usepackage[hang,flushmargin]{footmisc}
\usepackage{lipsum}
\DeclareMathOperator*{\argmin}{arg\,min}
\DeclareMathOperator*{\argmax}{arg\,max}
\DeclarePairedDelimiterX{\infdivx}[2]{(}{)}{%
  #1\;\delimsize\|\;#2%
}
\definecolor{nice-red}{HTML}{E41A1C}
\definecolor{nice-orange}{HTML}{FF7F00}
\definecolor{nice-yellow}{HTML}{FFC020}
\definecolor{nice-green}{HTML}{4DAF4A}
\definecolor{nice-blue}{HTML}{377EB8}
\definecolor{nice-nice-red}{HTML}{984EA3}
\definecolor{nice-grey}{HTML}{91A3B0}
\definecolor{nice-amber}{HTML}{DAA520}
\usepackage{algorithm}
\newcolumntype{P}[1]{>{\centering\arraybackslash}p{#1}}
\newcommand{\infdiv}{\infdivx}
\newcommand{\algorithmfootnote}[2][\footnotesize]{%
  \let\old@algocf@finish\@algocf@finish% Store algorithm finish macro
  \def\@algocf@finish{\old@algocf@finish% Update finish macro to insert "footnote"
    \leavevmode\rlap{\begin{minipage}{\linewidth}
    #1#2
    \end{minipage}}%
  }%
}
\allowdisplaybreaks
\title{\LARGE \bf
SQUIRL: Robust and Efficient Learning from Video Demonstration of Long-Horizon Robotic Manipulation Tasks}

\author{Bohan Wu, Feng Xu, Zhanpeng He, Abhi Gupta, and Peter K. Allen
\thanks{This work is supported by NSF Grant CMMI-1734557. Authors are with Columbia University Robotics Group, New York, NY, 10027, USA}
}
\begin{document}

\maketitle
\begin{abstract}
Recent advances in deep reinforcement learning (RL) have demonstrated its potential to learn complex robotic manipulation tasks. However, RL still requires the robot to collect a large amount of real-world experience. 
To address this problem, recent works have proposed learning from expert demonstrations (LfD), particularly via inverse reinforcement learning (IRL), given its ability to achieve robust performance with only a small number of expert demonstrations.
Nevertheless, deploying IRL on real robots is still challenging due to the large number of \textit{robot} experiences it requires.
This paper aims to address this scalability challenge with a robust, sample-efficient, and general meta-IRL algorithm, SQUIRL, that performs a new but related long-horizon task robustly given only a single video demonstration.
First, this algorithm bootstraps the learning of a task encoder and a task-conditioned policy using behavioral cloning (BC). It then collects real-robot experiences and bypasses reward learning by \textit{directly} recovering a Q-function from the combined robot and expert trajectories. Next, this algorithm uses the Q-function to re-evaluate all cumulative experiences collected by the robot to improve the policy quickly. In the end, the policy performs more robustly (90\%+ success) than BC on new tasks while requiring no trial-and-errors at test time.
Finally, our real-robot and simulated experiments demonstrate our algorithm's generality across different state spaces, action spaces, and vision-based manipulation tasks, e.g., pick-pour-place and pick-carry-drop.
\end{abstract}

\section{Introduction}
We aspire robots to become generalists who acquire new complex skills robustly and quickly. The robotic system, whether planned or learned, needs to leverage its existing knowledge to solve a new but related task in an efficient yet high-performance manner. 
Thanks to recent advances in machine learning and sim-to-real transfer mechanisms, short-horizon robotic manipulation such as grasping has improved in performance.
However, many real-world robotic manipulation tasks are long-horizon, diverse, and abundant in volume. In the absence of a scalable and systematic way to construct simulation environments for a large number of tasks, the robot needs to learn a new task directly in the physical world from only a handful of trials, due to the high cost of collecting real-robot trial-and-errors and experiences.

\begin{figure}
    \centering
    \includegraphics[width=1\linewidth]{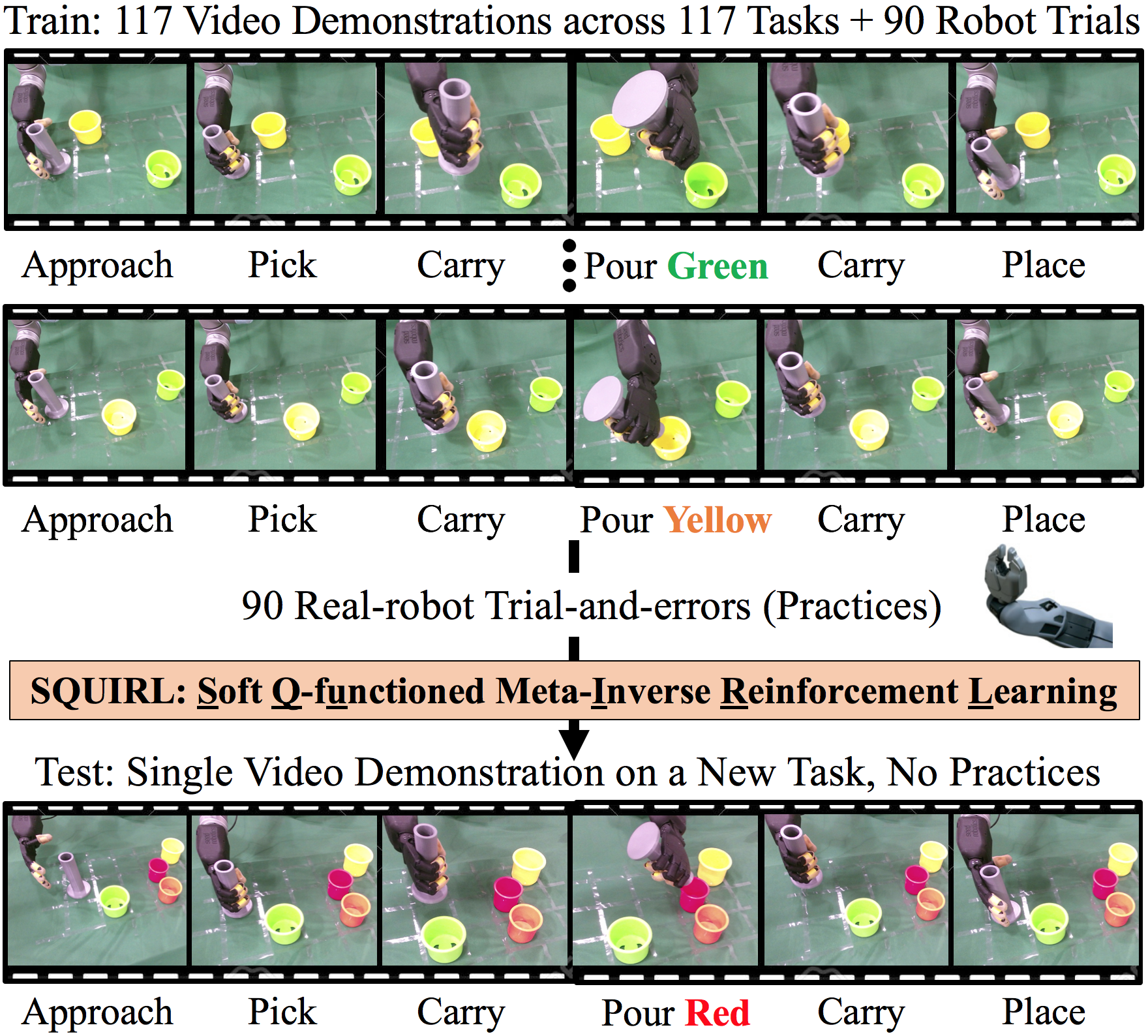}
    \caption{\small \textbf{Learning from a single video demonstration of a long-horizon manipulation task via Soft Q-functioned Meta-IRL (SQUIRL).} In the pick-pour-place example above, the robot needs to approach, pick-up and carry the grey bottle, pour the iron pebble inside the bottle into a specific container, and finally place the bottle back on the table. During training, the robot is given a single video demonstration for \textit{each} of the 117 training tasks. After learning from these 117 videos, the robot also practices 90 trial-and-errors \textit{in total} on these tasks. From such combined expert and robot trajectories, the robot learns the general skills of pouring robustly. At test time, given a single video demonstration of pouring into a \textit{new, unseen} red container at a \textit{new} position, the robot successfully replicates this new task without the need for any trial-and-errors.} 
    \label{fig:intro}
\end{figure}
We observe that real-world robotic skill acquisition can become more sample-efficient in several important ways. First, we notice that humans learn tasks quickly by watching others perform similar tasks. Among many forms of task representations such as rewards, goal images, and language instructions, human demonstrations guide exploration effectively and can lead to significant sample efficiency gains. 
Furthermore, learning from video demonstrations sidesteps hand-designing a proper reward function for every new task. In the case of a vision-based task, video demonstrations also conveniently provide the same pixel state space for the robot. 

In learning from demonstrations (LfD), the robot should be sample-efficient in two dimensions -- it should use as few expert demonstrations (``demonstrations'' hereafter) as possible and take as few trial-and-errors (practices) as possible on its own to learn a robust policy.
Among LfD methods, behavioral cloning (``BC'' hereafter) is sample-efficient but susceptible to compounding errors. 
Here, compounding errors refer to the problem in which every time a behavioral-cloned robot makes a small error, it makes a larger error down the road as it drifts away from the expert state distribution.
In contrast, IRL alleviates compounding errors by allowing the robot to try the tasks out in the real world and measure its behavior against the expert.
However, due to the need to learn a reward function, IRL can require many trial-and-errors in the real world, while BC does not require such robot experiences.
We posit that leveraging off-policy experiences of trial-and-errors is essential to making IRL sample-efficient enough for real robots. Here, ``off-policy experiences'' refer to the \textit{cumulative} experiences that the \textit{robot} has collected thus far during \textit{training}.
In contrast, ``on-policy experiences'' are the most recent experiences that the robot has collected using its \textit{current} policy. Humans leverage lifelong, cumulative experiences to learn quickly at present. We envision robots to acquire new skills more quickly by learning from off-policy (i.e., cumulative) experiences.

Finally, many real-world tasks are related and share structures and knowledge that can be exploited to solve a new but similar task later.
For example, humans can quickly learn to pick and place a new object after learning to pick and place many known objects.
Meta-learning, explicitly utilizing this property, aims to learn a new but related task quickly if it has already learned several similar tasks in the past.

\begin{SCfigure}[1][t]
\captionsetup{labelformat=empty}
\includegraphics[
trim={0 0 0 0},clip,
origin=c, width=0.5\linewidth]{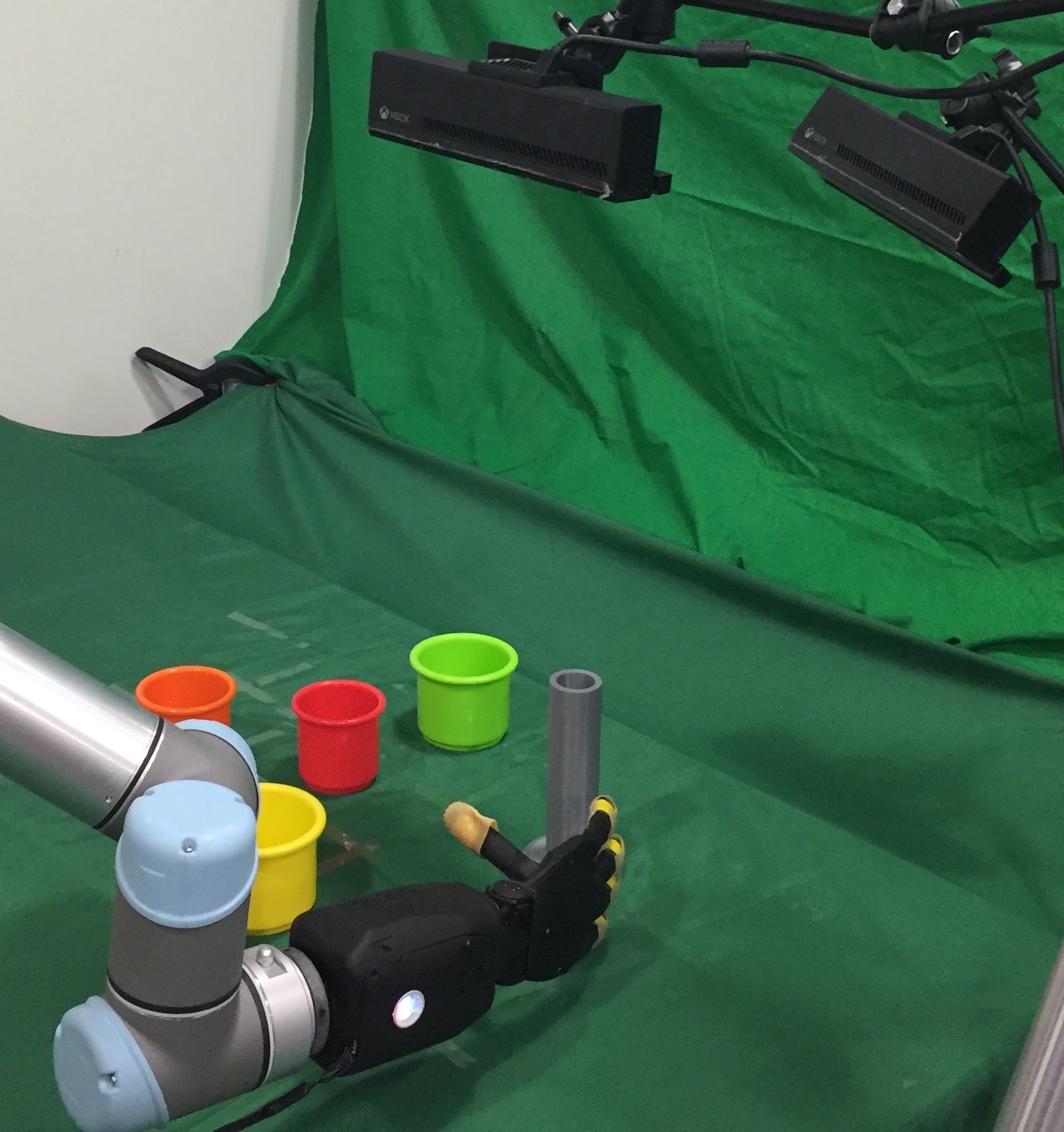}
\caption{\small Fig.2: \textbf{Pick-Pour-Place Robot Setup at Test Time}. Given an RGB image from the top-down (black) \textit{or} 45\degree camera (also black), the UR5-Seed robot is tasked to approach and pick-up the grey cylindrical bottle, pour the iron pebble already inside the bottle into a specific container on the table and finally place the bottle back on the table.}
\label{fig:hardware}
\end{SCfigure}
With these motivations, we introduce SQUIRL, a meta-IRL algorithm that learns long-horizon tasks quickly and robustly by learning from 1) video demonstrations, 2) off-policy robot experiences, and 3) a set of related tasks. Fig.\ref{fig:intro} explains this algorithm using the example of a set of long-horizon pick-pour-place tasks, using the UR5-Seed\footnote{Site:  \url{www.seedrobotics.com/rh8d-dexterous-hand.html}} robot setup shown in Fig.\ref{fig:hardware}. In this task, we have the containers (green, yellow, orange, and red), a cylindrical bottle (grey), and an iron pebble inside the bottle. The robot needs to \textit{first approach} and pick-up the grey bottle, pour the iron pebble inside the bottle into a specific container on the table, and then finally place the bottle back on the table, as shown in each row of images in Fig.\ref{fig:intro}. At the beginning of each task, the bottle is \textit{not yet} in hand, but the iron pebble is already in the bottle. At training time, the robot is given a single video demonstration for each of the 117 pick-pour-place tasks, as shown in the first two rows of images in Fig.\ref{fig:intro}. Every new combination of container positions represents a different pick-pour-place task. Furthermore, the robot only needs to pour into \textit{one} of the containers in a single task. Therefore, pouring into different containers also represents different tasks.
After learning from these 117 demonstrations, the robot also practices 90 trial-and-errors on these tasks \textit{in total}. From such a combination of expert and robot trajectories, the robot learns the general skills of pick-pour-place robustly. In all 117 training tasks, only \textit{two} of the four containers appear on the table: the green and yellow containers, as shown in the first two rows of images in Fig.\ref{fig:intro}. The orange and red containers are excluded during training and only appear at test time, as shown in the last row of images in Fig.\ref{fig:intro}. We do so to evaluate our algorithm's generalizability to unseen containers \textit{at test time}. As shown in the last row of images in Fig.\ref{fig:intro}, the robot successfully pours into a \textit{new} container (red) at test time, at a \textit{new} position never seen before during training, without the need for any trials or practices.

To achieve such fast generalization to new tasks, our algorithm learns a task encoder network and a task-conditioned policy. The task encoder generates a 32-dimensional task embedding vector that encodes task-specific information. The policy network then learns to generalize to new tasks by accepting this task embedding vector as input, thus becoming ``task-conditioned''. During training, our algorithm first bootstraps learning by training both the task encoder and the policy jointly via the BC loss. The robot then collects 10 trials across 10 tasks using the warmed-up policy and the task encoder. Next, using the combined experiences of the expert and the robot, our algorithm bypasses reward learning by directly learning a task-conditioned Q-function.
Using this Q-function, our algorithm then reuses and re-evaluates all cumulative experiences of the robot to improve the policy quickly. This cycle repeats until the $90^{th}$ trial.
Finally, at test time, the task encoder generates a new task embedding from a \textit{single} video demonstration of a new task. This embedding is then inputted into the task-conditioned policy to solve the new task without any trial-and-errors and yet in a high-performance manner. In summary, our contributions are:
\begin{enumerate}
    \item A \textit{robust} meta-IRL algorithm that outperforms ($90\%$+ success) its behavioral cloning counterpart in real-robot and simulated vision-based long-horizon manipulation;
    \item A \textit{novel} Q-functioned IRL formulation that circumvents reward learning and improves IRL sample efficiency;
    \item An \textit{efficient} method that leverages off-policy robot experiences for training and requires no trials at test time;
    \item A \textit{general} approach that tackles various long-horizon robotic manipulation tasks and works with both vision and non-vision observations and different action spaces.
\end{enumerate}

\section{Related Work}
\subsection{Inverse Reinforcement Learning (IRL) and Meta-IRL}
Inverse reinforcement learning (IRL) models another agent's (typically the expert's) reward function, given its policy or observed behavior.
Previous works have approached the IRL problem with maximum margin methods \cite{abbeel2004irl}\cite{ratliff2006mmp} and maximum entropy methods~\cite{ziebart2010entropyirl}\cite{ziebart2008maximum}\cite{boularias11a2011entropyirl}. 
In particular, maximum entropy methods recover a distribution of trajectories that have maximum entropy among all distributions and match the demonstrated policy's behaviors.
While these methods have shown promising results in continuous control problems, they suffer from low sample efficiency due to the need for evaluating the robot's policy, which can be alleviated by meta-learning (i.e., meta-IRL).
SMILe~\cite{smile} and PEMIRL~\cite{yu2019meta} are two meta-IRL algorithms based on AIRL~\cite{fu2018learning} that leverage a distribution of tasks to learn a continuous task-embedding space to encode task information and achieve fast adaptation to a new but similar task. Our work differs from~\cite{smile}\cite{yu2019meta} in four crucial ways. First, our meta-IRL algorithm works with real robots and image observations. Second, instead of a reward function, we directly model a Q-function that the policy can optimize, in order to increase IRL sample efficiency. Third, we train the task encoder with the behavioral cloning (BC) gradient as opposed to the IRL gradient for stabler and more efficient learning. Lastly, we bootstrap policy and task encoder learning using BC before training via meta-IRL. 

\subsection{Real-robot Learning from Demonstrations (LfD)}
Our work is related to real-robot LfD \cite{argall-survey-robot-learning-2009}, such as~\cite{xu2018neural}\cite{huang2019neural}\cite{huang2019motion}.
In particular, \cite{finn2016guided} developed IRL on real robots without learning from raw pixels. Other works (e.g., \cite{zhang2017imitationmanipulation}\cite{Kober2010RAM}\cite{paster2009motorskills}\cite{sermanet2018time}) used BC for real-robot LfD.
Another work \cite{lynch2019latentplan} developed goal-conditioned BC on a simulated robot to learn long-horizon tasks by playing with objects in the scene.
While enjoying efficient learning by casting imitation learning into a supervised learning problem, BC suffers from the covariate shift between the train and test data.
In comparison, IRL achieves robust performance by modeling the state-action joint distribution instead of the conditional action distribution in BC~\cite{divergence2019}.
Different from previous works, our meta-IRL algorithm works on real-robot \textit{vision-based} tasks, and its Q-functioned IRL policy gradient can be directly combined with the BC gradient signal to approach both the sample efficiency of BC and the robustness of IRL.

\subsection{One-shot Meta-imitation Learning on Real Robots}
Our algorithm attempts to enable robots to quickly and robustly imitate a single unseen video demonstration by learning from a distribution of tasks with shared structure, i.e., one-shot robot meta-imitation learning.
For example, \cite{finn2017one} combines gradient-based meta-learning and BC on a real robot to learn quickly from video demonstrations. \cite{yu2018one} then extends \cite{finn2017one} to enable robots to learn from human-arm demonstrations directly. 
\cite{yu2019one} then improves~\cite{yu2018one} to meta-imitation-learn multi-stage real-robot visuomotor tasks in a hierarchical manner.
However, constrained by the covariate shift problem of BC, these works show limited task performance (e.g., around $50\%$ success rate for the training tasks). In contrast, our algorithm learns a vision-based manipulation task robustly ($90\%+$ success rates) and efficiently (117 videos, 90 trials) by utilizing the generalization ability of task embeddings~\cite{rakelly2019efficient} and a novel Q-functioned IRL formulation.

\section{Preliminaries}
\subsection{Off-policy Reinforcement Learning via Soft Actor-Critic}
Standard RL models a task $\mathcal{M}$ as an MDP defined by a state space $\mathcal{S}$, an initial state distribution $\rho_0 \in \Pi(\mathcal{S})$, an action space $\mathcal{A}$, a reward function $\mathcal{R}: \mathcal{S} \times \mathcal{A} \to \mathbb{R}$, a dynamics model $\mathcal{T}: \mathcal{S} \times \mathcal{A} \to \Pi(\mathcal{S})$, a discount factor $\gamma \in [0, 1)$, and a horizon $H$.
Here, $\Pi(\cdot)$ defines a probability distribution over a set.
The robot acts according to stochastic policy $\pi: \mathcal{S} \to \Pi(\mathcal{A})$, which specifies action probabilities for each $s$.
Each policy $\pi$ has a corresponding $Q^\pi: \mathcal{S}\times\mathcal{A} \to \mathbb{R}$ function that defines the expected discounted cumulative reward for taking an action $a$ from $s$ and following $\pi$ onward. 

Off-policy RL, particularly Soft Actor-Critic (SAC)~\cite{haarnoja2018soft}, reuses historical experiences to improve learning sample efficiency by recovering a ``soft'' Q-function estimator $Q_{\theta}$. A policy can then be learned by minimizing the KL divergence between the policy distribution and the exponential-Q distribution: $\pi^* = \argmin_{\pi \in \Pi} D_{KL} \infdiv{\pi(a \mid s)}{\frac{\exp(Q^{\pi_{old}}_{\theta}(s, a))}{Z(s)}}$

\subsection{Timestep-centric IRL as Adversarial Imitation Learning}
\label{sec:Timestep-centric}
The purpose of IRL is to learn the energy function $f_\theta$ implicit in the provided expert demonstrations and use $f_\theta$ to learn a policy that robustly matches the expert performance.
In particular, timestep-centric IRL aims to recover an energy function $f_\theta(s, a)$ to rationalize and match the demonstrated expert's action conditional distribution: $p_{\theta}(a \mid s) = \frac{\exp(f_\theta(s, a))}{Z_\theta} \propto \exp(f_\theta(s, a)) = \overline{p_{\theta}}(a \mid s)$,
where $Z_\theta$ is the partition function, an integral over all possible actions given state $s$. In other words, IRL minimizes the KL divergence between the actual and predicted expert conditional action distributions: $\pi_E(a \mid s)$ and $p_{\theta}(a \mid s)$.

Adversarial IRL~\cite{fu2018learning}\cite{ho2016generative} provides a sampling-based approximation to MatEntIRL~\cite{ziebart2008maximum} in an adversarial manner. Specifically, AIRL~\cite{fu2018learning} learns a generative policy $\pi_\psi$ and a binary discriminator $D_\theta$ derived from energy function $f_\theta$:
\begin{align}
\label{eq:dtheta}
    &D_\theta(s, a) = P((s, a) \text{ is generated by expert})\nonumber\\
    &= \frac{\overline{p_{\theta}}(a \mid s) }{\overline{p_{\theta}}(a \mid s)  + \pi_\psi(a \mid s)} = \frac{\exp (f_\theta(s, a))}{\exp (f_\theta(s, a)) + \pi_\psi(a \mid s)}
\end{align}
and $\theta$ is trained to distinguish state-action pairs sampled from the expert vs. the policy, using binary cross entropy loss:
\begin{align}
\label{eq:discriminatorloss}
    \mathcal{L}^{IRL} = -\mathbb{E}&_{(s, a) \sim \pi_\psi, \pi_E}[y(s, a) \log(D_\theta(s, a)) \nonumber\\
    &+ (1-y(s, a))\log(1-D_\theta(s, a))]
\end{align}
where $y(s, a) = \mathds{1}\{(s, a) \text{ is generated by expert }\pi_E\}$.

Meanwhile, the policy is trained to maximize the MaxEntIRL Objective~\cite{ziebart2008maximum}, or equivalently, to match the expert's state-action joint distribution via reverse KL divergence~\cite{divergence2019}.
\subsection{One-shot Meta-imitation Learning from A Single Video}
In one-shot meta-imitation learning, the robot is trained to solve a large number of tasks drawn from a task distribution $p(\mathcal{M})$.
The total number of tasks in this task distribution can be finite or infinite. Each imitation task $\mathcal{M}_{train}^i$ consists of a single video demonstration $\mathcal{D}^i_{\pi_E}$. During training, the robot can also generate limited practice trajectories (e.g., 90). For example, in the Pick-Pour-Place experiment in Fig.\ref{fig:intro}, the robot receives a single video demonstration for each of the 117 tasks. Each task is characterized by a different combination of container positions, or pouring into the green vs. the yellow container. At test time, the robot receives a single video of a new task $\mathcal{M}_{test}^i$ drawn from $p(\mathcal{M})$. For example, a new Pick-Pour-Place test task can be a \textit{new} combination of container positions or pouring into a \textit{new} container (e.g., the red or orange container). The robot then needs to solve this task the first time \textit{without trial-and-error}. 
\subsection{Embedding-based Meta-learning}
Embedding-based meta-learning~\cite{yu2019meta}\cite{rakelly2019efficient} learns a task-specific embedding vector $z$ that contains task-level abstraction to adapt to a new but related task quickly. This method aims to learn a task-conditioned policy $\pi(a | s, z)$ that maximizes task-conditioned expected returns:
$\max_\pi  \mathbb{E}_{(s_t, a_t) \sim \pi, \rho_0} [\sum_{t=1}^T r(s_t, a_t | c) + \alpha \mathcal{H}(\pi(a_t|s_t, c))]$, by learning an embedding space $Z$ that maximizes the mutual information between $z$ and task context $c$. The goal is to make this learned embedding space generalizable to new tasks so that at test time, the policy can quickly adapt to unseen tasks with no or few practices. A key advantage of embedding-based meta-learning is the ability to learn from off-policy experiences. However, current methods are mostly if not only demonstrated in non-vision tasks in simulation. 

\section{Mathematical Formulation for SQUIRL}
\subsection{SQUIRL: Timestep-centric IRL as Soft Q-Learning}
\label{sec:q}
Previous works in timestep-centric IRL such as~\cite{smile}\cite{yu2019meta}\cite{fu2018learning} have interpreted the energy function $f_\theta$ in Eq.\ref{eq:dtheta} as a reward function $r_\theta$ and later recover a Q or advantage function based on reward $r_\theta$ for policy improvement. To improve IRL sample efficiency, we propose to \textit{bypass} this reward learning and directly interpret $f_\theta(s,a)$ as the soft Q-function~\cite{haarnoja2018soft} $Q^{\pi_{mix}}_\theta(s,a)$. This soft Q-function models the expert's behavior as maximizing both the Q-value and its \textit{entropy} (i.e., randomness) simultaneously. It also encourages the robot to explore the real world to imitate the expert more robustly. Under this formulation, approximating the expert's conditional action distribution is equivalent to recovering a soft Q-function under which the expert is soft Q-optimal:
\begin{align}
&\argmin_\theta D_{KL}\infdiv{\pi_E(a \mid s)}{p_{\theta}(a \mid s)} \nonumber\\
= &\argmax_\theta \mathbb{E}_{a \sim \pi_E(a \mid s)} [Q^{\pi_{mix}}_\theta(s, a)] - \log Z_\theta 
\label{eq:softqtheta}
\end{align}
Eq.\ref{eq:softqtheta} rationalizes the expert behavior intuitively because the expert should be optimal with respect to the \textit{cumulative} reward~\cite{ziebart2010entropyirl}, not the immediate reward. Here, $Q^{\pi_{mix}}_\theta$ is under a mixture policy $\pi_{mix}$ between the robot and expert's policies.

\subsection{SQUIRL as Expert Imitation and Adversarial Learning}
\label{sec:match}

Under SQUIRL, the policy learning objective (Eq.\ref{eq:rl}) is also equivalent (derivations on website) to matching: 1) the exponential-Q distribution of the discriminator $\theta$ (Eq.\ref{eq:match}), 2) the generator's objective in Generative Adversarial Networks (GANs)~\cite{goodfellow2014generative} (Eq.\ref{eq:gan}), and 3) the joint state-action distribution of expert~\cite{divergence2019} (Eq.\ref{eq:joint}): $\pi^* = \argmin_{\pi \in \Pi} \mathcal{L}^{RL}(\pi)$, where
\begin{align}
    \label{eq:rl}&\mathcal{L}^{RL}(\pi) =D_{KL} \infdiv{\pi_\psi(a \mid s)}{\frac{\exp{Q^{\pi_{mix}}_{\theta}(s, a)}}{Z(s)}} \\
     \label{eq:match}&= D_{KL} \infdiv{\pi_\psi(a \mid s)}{p_\theta(a \mid s)}\\
 \label{eq:gan}&= \mathbb{E}_{(s,a) \sim \pi_{mix}}[\log(1-D_\theta(s,a))-\log(D_\theta(s,a))]\\
 \label{eq:joint}&=  D_{KL} \infdiv{\rho_{\pi_\psi}(s, a)}{\rho_{\pi_E}(s, a)}
\end{align}

Meanwhile, the discriminator $\theta$ is matching its Q-function to the log-distribution of the expert's conditional action distribution (Section~\ref{sec:Timestep-centric}). Therefore, when this Q-function is optimal: $Q^{\pi_{mix}}_\theta = Q^{\pi_{mix}}_{\theta^*}$, the robot's policy objective (Eq.\ref{eq:rl}) is also matching the expert's conditional action distribution: 
\begin{equation}
\label{eq:indirect}
\psi^* = \argmin_\psi E_{\rho_{\pi_{mix}}(s)} [D_{KL} \infdiv{\pi_\psi(a \mid s)}{\pi_E(a \mid s)}]
\end{equation}

\subsection{Comparison to the Behavioral Cloning (BC) Objective}
\label{sec:bc}
While BC attempts to learn a policy that also matches the expert's conditional action distribution~\cite{divergence2019}, the \textit{fundamental} difference is that the KL-divergence in BC's case is computed under the \textit{expert's} narrow state distribution $\rho_{\pi_E}(s)$: $\psi_{BC}^* = \argmin_\psi E_{\rho_{\pi_{E}}(s)} [D_{KL} \infdiv{\pi_E(a \mid s)}{\pi_\psi(a \mid s)}]$. In contrast, ours (Eq.\ref{eq:indirect}) is computed under $\rho_{\pi_{mix}}(s)$: the state distribution of the \textit{combined} \textit{cumulative} experience of the robot and the expert, which is a much wider distribution than the expert distribution. We hypothesize that this, along with matching the joint state-action distribution of the expert (Eq.\ref{eq:joint}), makes our algorithm less susceptible to compounding errors than BC, as experimentally tested in Section~\ref{sec:exp}. 

\begin{figure*}
    \centering
    \includegraphics[width=\linewidth]{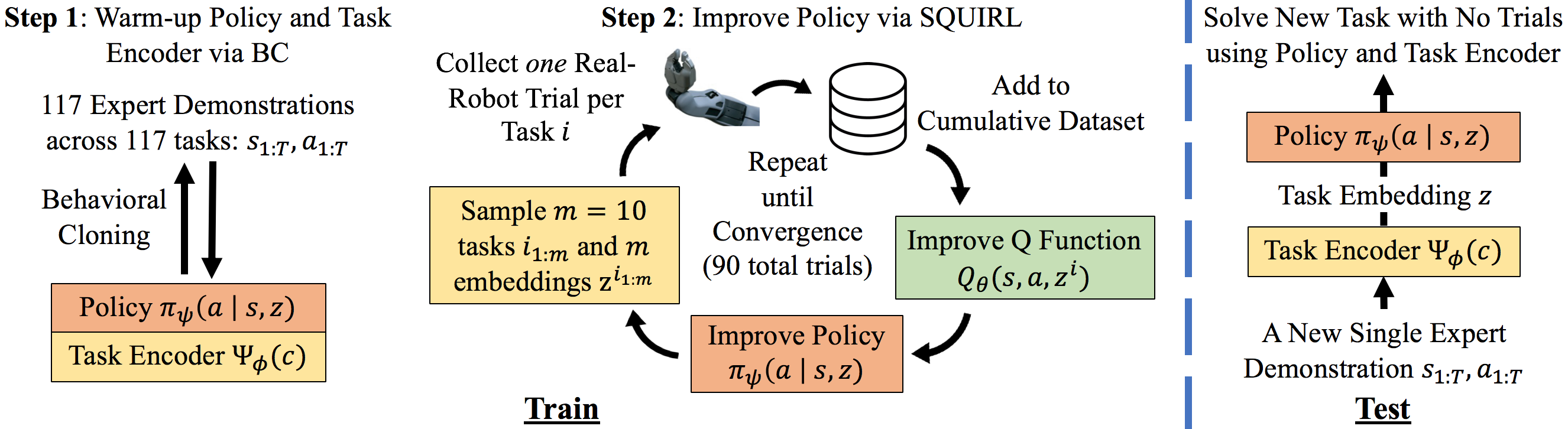}
    \caption{\small \textbf{SQUIRL: Soft Q-functioned Meta-IRL}. To begin, our algorithm bootstraps learning for the policy (orange) and the task encoder (yellow) via behavioral cloning (the left third of Fig.3). Next, our algorithm uses the warmed-up policy and task encoder to generate 10 trials \textit{in the physical world} (not in simulation). Using the combined expert and robot trajectories, our algorithm learns a task-conditioned soft Q-function (green) that rationalizes the expert's behaviors as maximizing both cumulative reward and entropy (i.e., randomness). Using this Q-function, our algorithm then quickly improves the policy using \textit{all cumulative} robot and expert timesteps. This cycle repeats until convergence, totaling \textit{90 trials} (the middle third of Fig.3). Finally, at test time (the right third Fig.3), our algorithm generates a new embedding $z$ for the new task, and inputs this embedding into the task-conditioned policy to solve the new task without any practices.\vspace{-.3cm}}
    \label{fig:architecture}
\end{figure*}
\section{SQUIRL: Soft Q-functioned Meta-IRL}
Shown in Fig.\ref{fig:architecture}, our algorithm learns three neural networks jointly -- a task encoder (yellow), a task-conditioned policy (orange), and a task-conditioned soft Q-function (green):
\begin{enumerate}
    \item $\Psi_{\phi} (c)$: a \textbf{task encoder} that encodes a sampled batch of $C=64$ expert state-action pairs $c = \{s^i_{1:C}, a^i_{1:C}\}$ from a task $i$ into a single 32-dim embedding vector $z^i \in\mathbb{R}^{32}$ (by computing the mean vector across 64 embeddings) that enables generalization to new tasks. This batch of expert state-action pairs is randomly sampled and thus does not encode time information. Both the policy and the Q-function accept this embedding vector as input. 
    \item $\pi_\psi(s, z^i)$: a \textbf{task-conditioned policy} the robot uses to perform a task $i$ given state $s$ and the task embedding vector $z^i \in \mathbb{R}^{32}$ outputted by the task encoder $\Psi_{\phi}(c)$.
    \item  $Q_\theta(s,a,z^i)$: a \textbf{task-conditioned soft Q-function} used to train the policy $\pi_\psi(s, z^i)$ to more robustly mimic the expert's behavior for the robotic manipulation task $i$.
\end{enumerate}

To begin, the robot is given an expert trajectory of state-action pairs $\mathcal{D}_{\pi_E}$ for each of the 117 training tasks. The robot first uses these expert trajectories to bootstrap training for both its policy $\pi_\psi$, and the task encoder $\Psi_\phi$ via behavioral cloning (Eq.\ref{eq:bc}). This way, the robot can distinguish the train tasks better and learn more quickly in the real world. Next, the robot generates 10 trials (state-action pairs) $\overline{\mathcal{D}}_{\pi_\psi}$ \textit{in the physical world} (not simulation) using its warmed-up policy and task encoder. Then, the robot uses both the expert's and its state-action pairs to train a discriminator $\theta$. This discriminator classifies which state-action pairs come from the expert $\pi_E$  vs. the robot $\pi_\psi$. At first, the robot is distinctively worse than the expert at performing the tasks. This makes it easy for the discriminator to classify. By doing so, the discriminator learns a Q-function $Q^{\pi_{mix}}_\theta$ using Eq.\ref{eq:softqtheta}. 

Using the learned Q-function $Q^{\pi_{mix}}_\theta$, the robot trains its policy $\pi_\psi$ via Eq.\ref{eq:rl}. Meanwhile, the robot also has the option to continue updating its task-conditioned policy and task encoder via behavioral cloning~(Eq.\ref{eq:bc}). Since training the policy via Eq.\ref{eq:rl} is equivalent to indirectly imitating the expert (Eq.\ref{eq:joint} and~\ref{eq:indirect}), as derived in Section~\ref{sec:match}, the trajectories generated by the policy gradually become more similar to the expert. This makes the state-action pairs more difficult for the discriminator to classify. This difficulty, in turn, forces the discriminator to learn a more precise Q-function, which then encourages the policy to mimic the expert even more closely. This cycle repeats until convergence (90 trials \textit{in total}), at which point: 1) the policy matches the expert performance, 2) the task encoder learns to generalize to new tasks, and 3) the discriminator continues to struggle to distinguish state-action pairs correctly despite having learned an accurate Q-function.

\subsection{Rationale for Bypassing Reward Learning via SQUIRL} SQUIRL learns a Q-function without rewards because 1) the policy is ultimately trained by the Q-function, not rewards, thus bypassing reward learning improves IRL sample efficiency, and 2) circumventing reward learning avoids off-policy Q-learning from a \textit{constantly changing} reward function and makes training easier and more stable empirically.

\subsection{Architectures for Policy, Task Encoder, and Q-function}
For all non-vision tasks, we parameterize $\pi_\psi, \Psi_\phi, Q_\theta$ with five fully-connected (FC) layers. For vision tasks, we use a 5-layer CNN followed by a spatial-softmax activation layer for the RGB image. This activation vector is then concatenated with the non-vision input vector and together passed through five FC layers. Our algorithm is \textit{general} and works with many other network architectures, state, and action spaces. 

\subsection{Incorporating BC to Bootstrap and Accelerate Learning}
Since our algorithm's IRL objective~(Eq.\ref{eq:indirect}) is compatible with BC, as explained in Section~\ref{sec:bc}, our algorithm can jointly be trained with BC to stabilize and accelerate learning without conflicting gradient issues (line 16 in Algorithm~\ref{algo:irl}): 
\begin{equation}
    \mathcal{L}^{BC} = \mathbb{E}_{(s, a) \sim \pi_E} [\left\lVert\pi_\psi (s, \Psi_\phi(c)) - a\right\rVert^2]
\label{eq:bc}
\end{equation}
This, combined with the off-policy nature of our algorithm, also allows the robot to bootstrap learning by first ``pre-training'' via BC (Eq.\ref{eq:bc}) using the expert demonstrations, before improving performance further via meta-IRL training.

\begin{algorithm}
\caption{SQUIRL: Soft Q-functioned Meta-IRL (Train)}
\hspace*{\algorithmicindent} \textbf{Input:} One expert video demonstration trajectory of state-action pairs $\mathcal{D}^i_{\pi_E}=\{s^i_{1:H}, a^i_{1:H}\}$ for each of the $n$ training tasks $i = 1:n$, where $H$ is the horizon of the task (e.g., $n=117, H=100$)
\begin{algorithmic}[1]
  \STATE Initialize soft Q-function $Q_\theta$, policy $\pi_\psi$, task encoder $\Psi_{\phi}$, and an empty buffer of off-policy robot trajectories $\mathcal{D}^i_{\pi_\psi} \leftarrow \{\}$ for each training task $i = 1:n$
  \STATE Warm-up policy and task encoder via $\mathcal{L}^{BC}$ (Eq.\ref{eq:bc})
  \WHILE{not converged}
  \STATE Sample a batch of $m$ task indices $\{i^{1:m}\}$ from all training tasks $i=1:n$, (e.g., $m=10$)
  \FOR{$i = i^{1:m}$}
  \STATE Infer task embedding $z^i=\mathbb{R}^\mathcal{Z} \leftarrow \Psi_\phi(c)$, where $c = \{s^i_{1:C}, a^i_{1:C}\} \sim \mathcal{D}^i_{\pi_E}$ (e.g., $\mathcal{Z} = 32, C=64$)
  \STATE Generate a robot trajectory of state-action pairs $\overline{\mathcal{D}}^i_{\pi_\psi} = \{s^i_{1:H}, a^i_{1:H}\}$ from task $i$ using $\pi_\psi, z^i$
  \STATE $\mathcal{D}^i_{\pi_\psi} \leftarrow \mathcal{D}^i_{\pi_\psi} \cup \overline{\mathcal{D}}^i_{\pi_\psi}$
  \ENDFOR
  \FOR{$j=1 : J$ (e.g., $J=400$)}
    \STATE Sample another batch of $m$ task indices $\{i^{1:m}\}$
    \STATE $\theta \leftarrow \theta - \nabla_\theta \mathcal{L}^{IRL}$ (Eq.\ref{eq:discriminatorloss}) using a combined batch of $\mathcal{B}=128$ robot and expert timesteps: $\overline{\mathcal{D}}^i_{\pi_\psi} \cup \overline{\mathcal{D}}^i_{\pi_E}$ and $z^i$, where $\overline{\mathcal{D}}^i_{\pi_\psi} \sim \mathcal{D}^i_{\pi_\psi}$, $\overline{\mathcal{D}}^i_{\pi_E} \sim \mathcal{D}^i_{\pi_E}$, $i=\{i^{1:m}\}$ 
    \ENDFOR
    \FOR{$k=1 : K$ (e.g., $K=2000$)}
    \STATE Sample another batch of $m$ task indices $\{i^{1:m}\}$
    \STATE{\textbf{if} necessary \textbf{then}} $\{\psi, \phi\} \leftarrow \{\psi, \phi\} - \nabla_{\psi, \phi} \mathcal{L}^{BC}$ (Eq.\ref{eq:bc}) using a batch of $\mathcal{B}$ expert timesteps  $\overline{\mathcal{D}}^i_{\pi_E} \sim \mathcal{D}^i_{\pi_E}, z^i$, $i=\{i^{1:m}\}$ \textbf{end if}
    \STATE $\psi \leftarrow \psi - \nabla_\psi \mathcal{L}^{RL}$ (Eq.\ref{eq:rl}) using a combined batch of $\mathcal{B}$ robot and expert timesteps: $\overline{\mathcal{D}}^i_{\pi_\psi} \cup \overline{\mathcal{D}}^i_{\pi_E}$ and $z^i$, where $\overline{\mathcal{D}}^i_{\pi_\psi} \sim \mathcal{D}^i_{\pi_\psi}$, $\overline{\mathcal{D}}^i_{\pi_E} \sim \mathcal{D}^i_{\pi_E}$, $i=\{i^{1:m}\}$ 
  \ENDFOR
  \ENDWHILE
  \STATE \textbf{return} soft Q-function $Q_\theta$, policy $\pi_\psi$, task encoder $\Psi_{\phi}$
\end{algorithmic}
\label{algo:irl}
\end{algorithm}
\begin{algorithm}
\caption{SQUIRL: Soft Q-functioned Meta-IRL (Test)}
\hspace*{\algorithmicindent} \textbf{Input:} $\pi_\psi$, $\Psi_\phi$, $Q_\theta$, and a single expert video demonstration of state-action pairs $\mathcal{D}^i_{\pi_E}=\{s^i_{1:H}$, $a^i_{1:H}\}$ from a \textit{new} task $i$ unseen during training
\begin{algorithmic}[1]
\STATE Infer task embedding vector $z^i=\mathbb{R}^\mathcal{Z} \leftarrow \Psi_\phi(c)$, where $c = \{s^i_{1:C}, a^i_{1:C}\} \sim \mathcal{D}^i_{\pi_E}$ (e.g., $\mathcal{Z} = 32, C=64$)
\STATE Rollout robot trajectory in the real world using $\pi_\psi$, $z^i$
  \end{algorithmic}
\label{algo:irltest}
\end{algorithm}
\begin{figure*}[h]
\captionsetup{justification=centering}
  \centering
    \minipage{0.162\textwidth}\includegraphics[trim={600 900 600 550},clip,width=\linewidth]{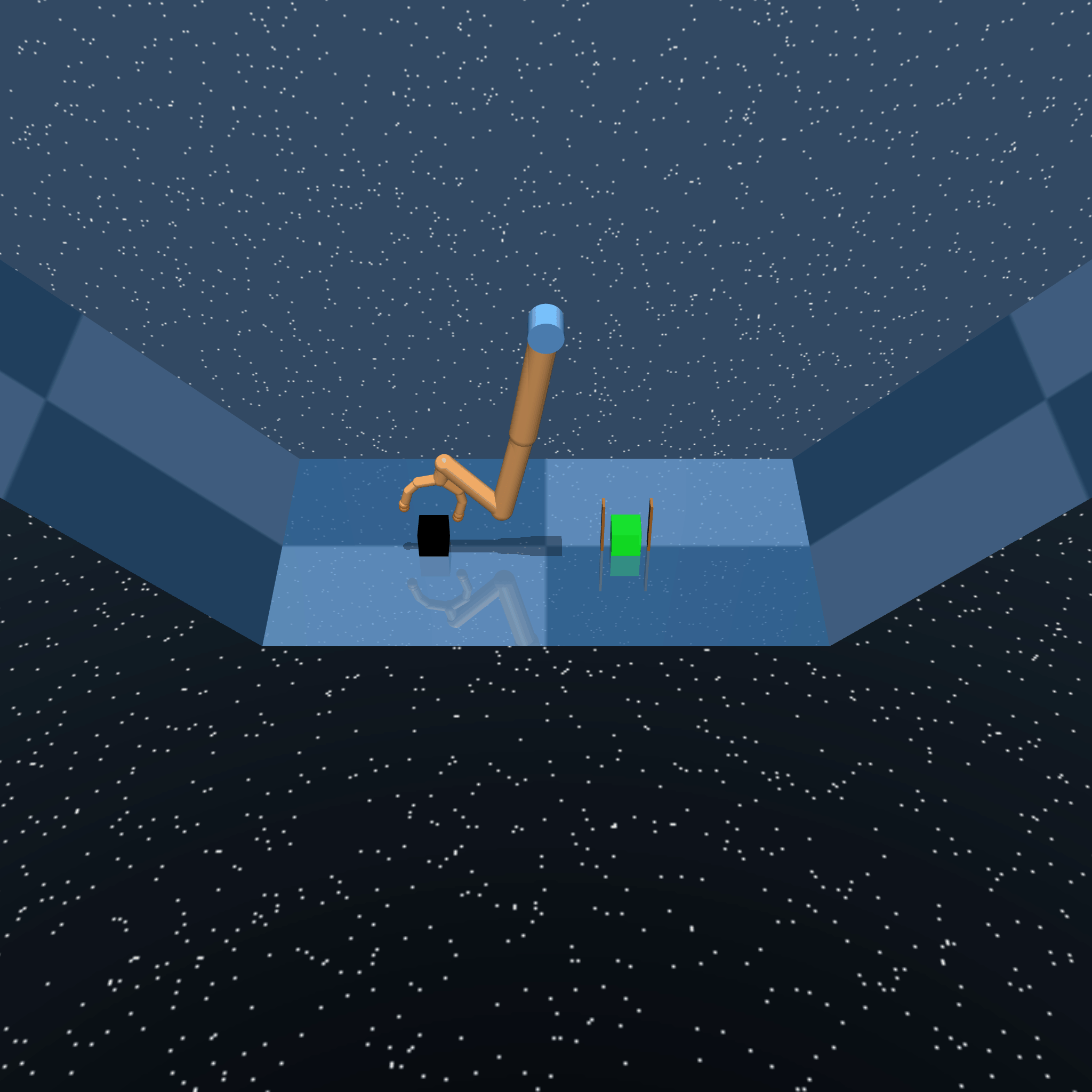}{\caption*{\small Approach Box}}\endminipage\hfill
    \minipage{0.162\textwidth}\includegraphics[trim={600 900 600 550},clip,width=\linewidth]{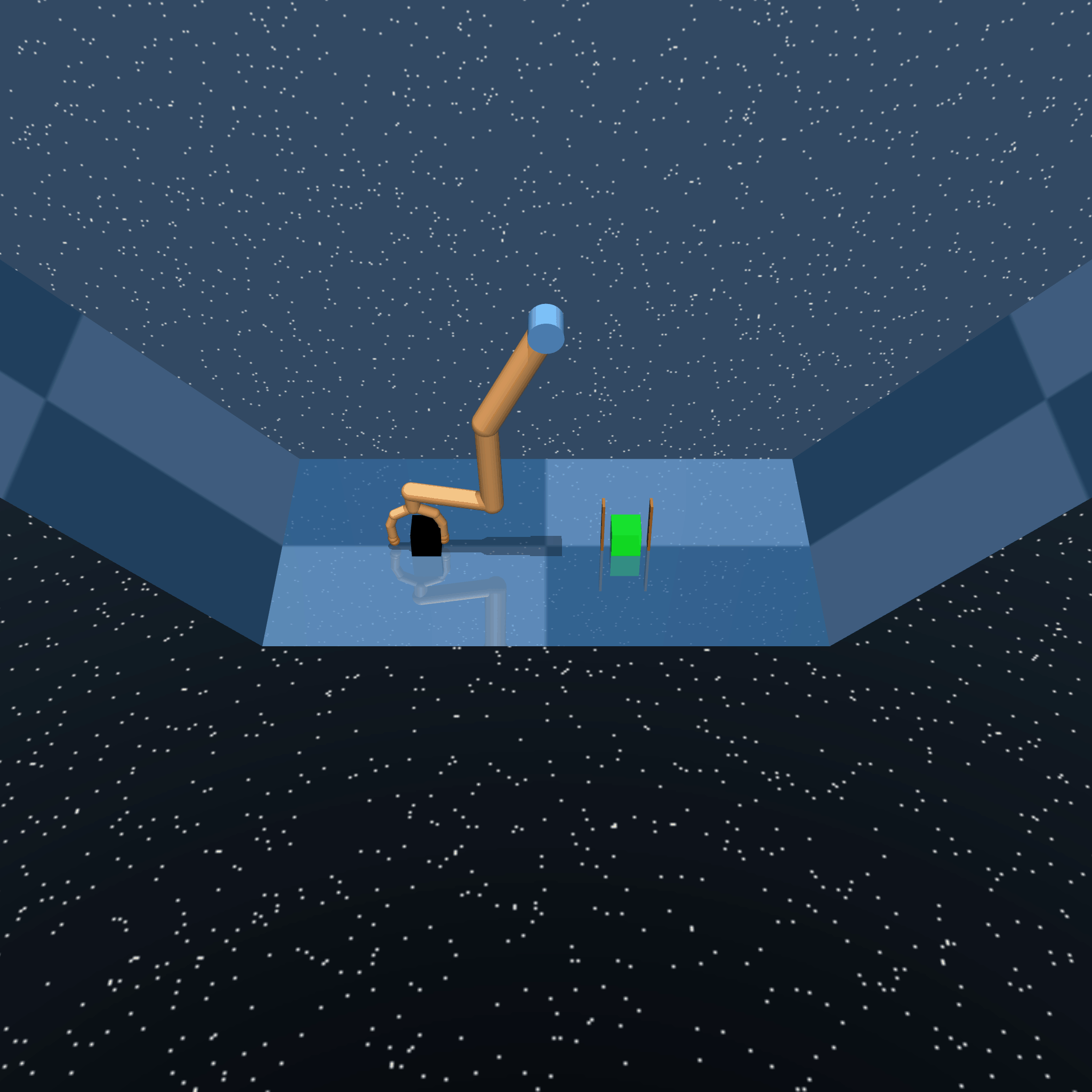}{\caption*{\small Lower to Box}}\endminipage\hfill
    \minipage{0.162\textwidth}\includegraphics[trim={600 900 600 550},clip,width=\linewidth]{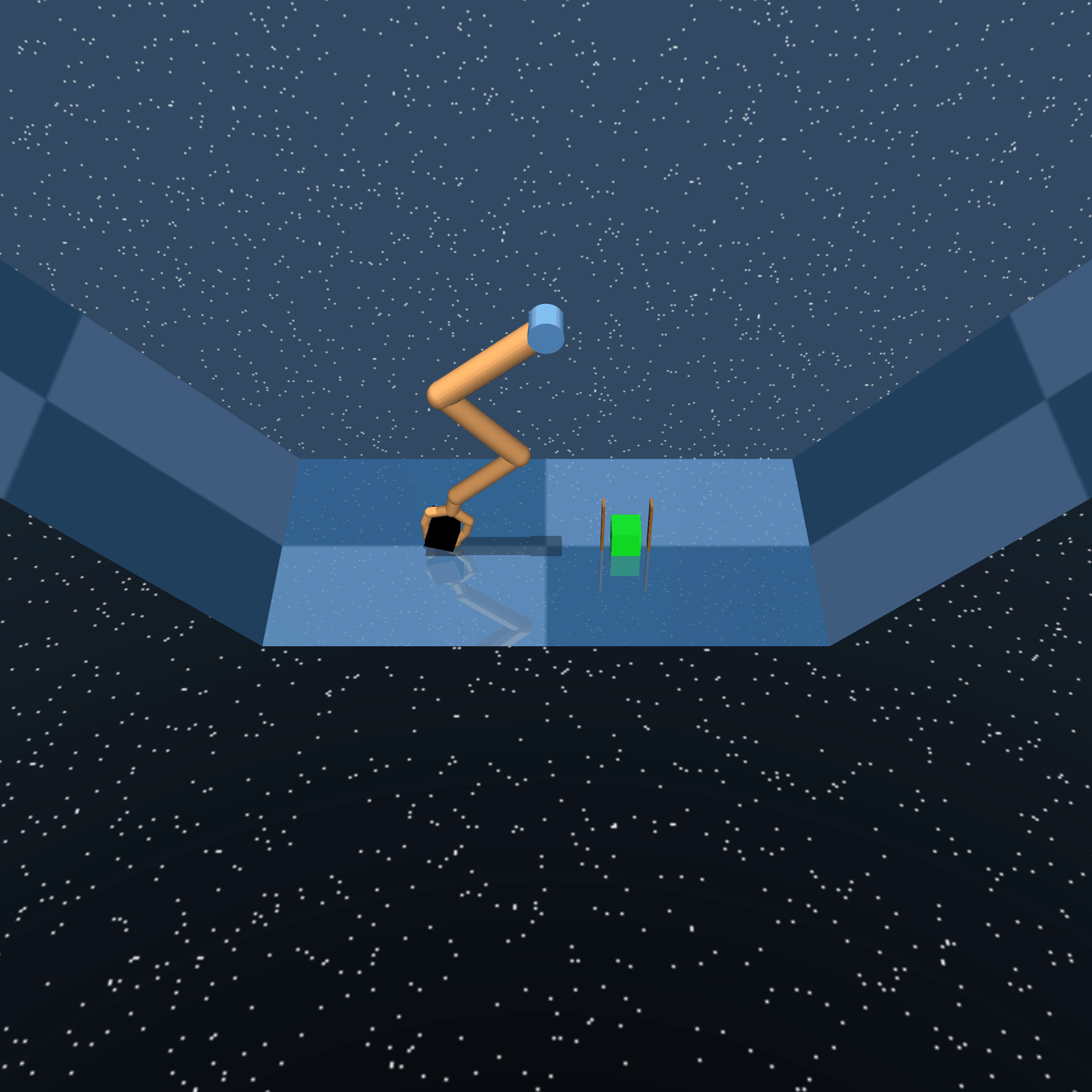}{\caption*{\small Grasp Box}}\endminipage\hfill
    \minipage{0.162\textwidth}\includegraphics[trim={600 900 600 550},clip,width=\linewidth]{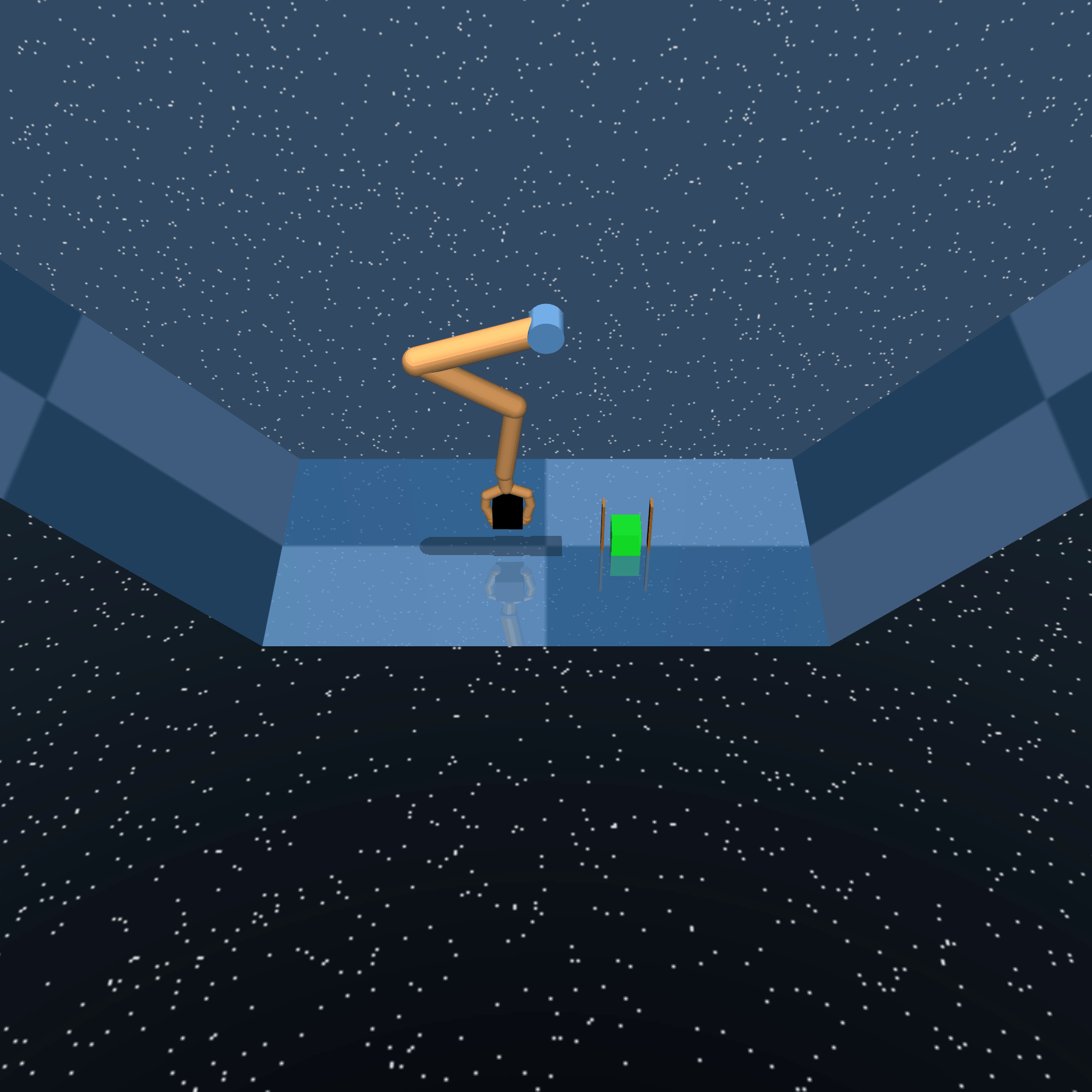}{\caption*{\small Pick up Box}}\endminipage\hfill
    \minipage{0.162\textwidth}\includegraphics[trim={600 900 600 550},clip,width=\linewidth]{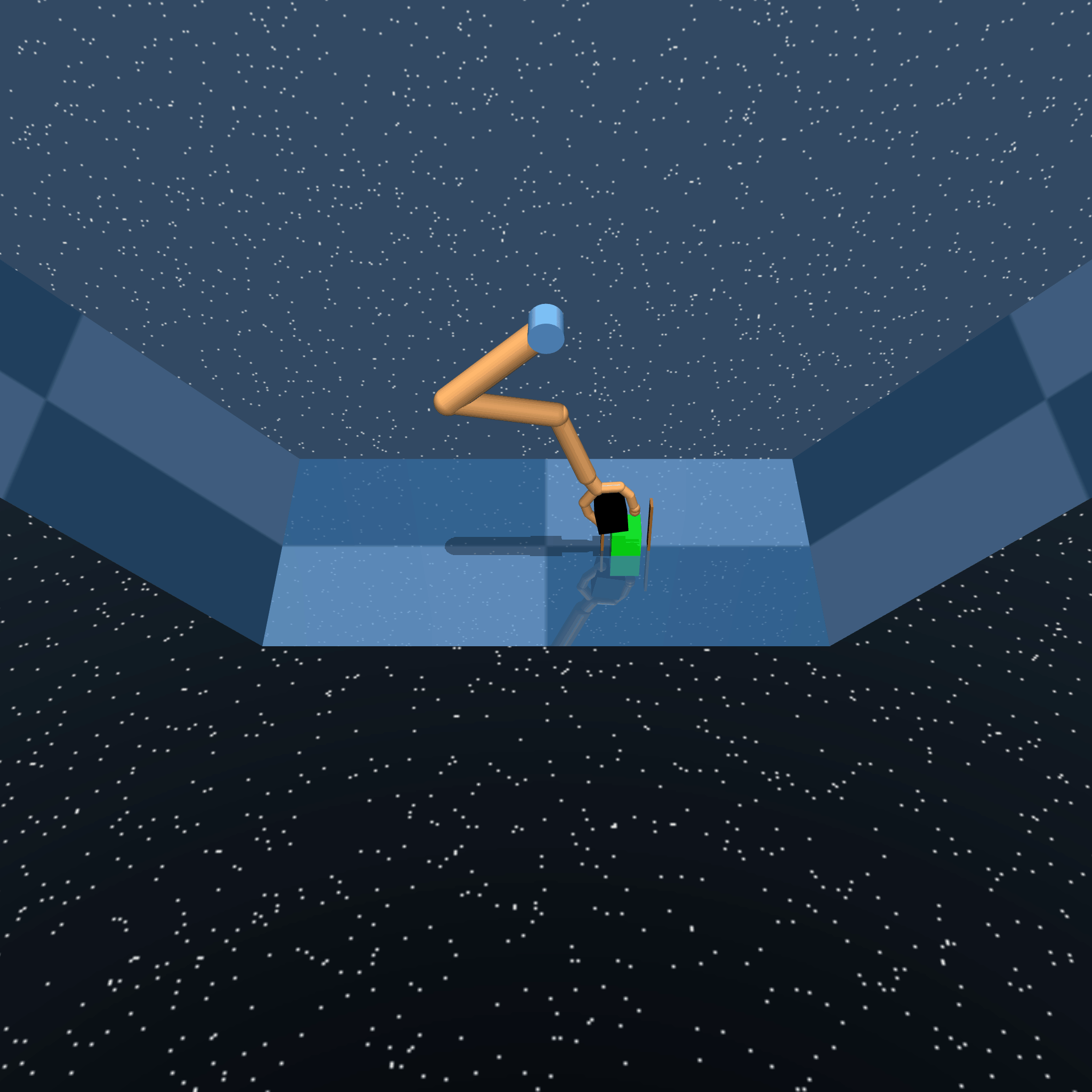}{\caption*{\small Carry Box}}\endminipage\hfill
    \minipage{0.162\textwidth}\includegraphics[trim={600 900 600 550},clip,width=\linewidth]{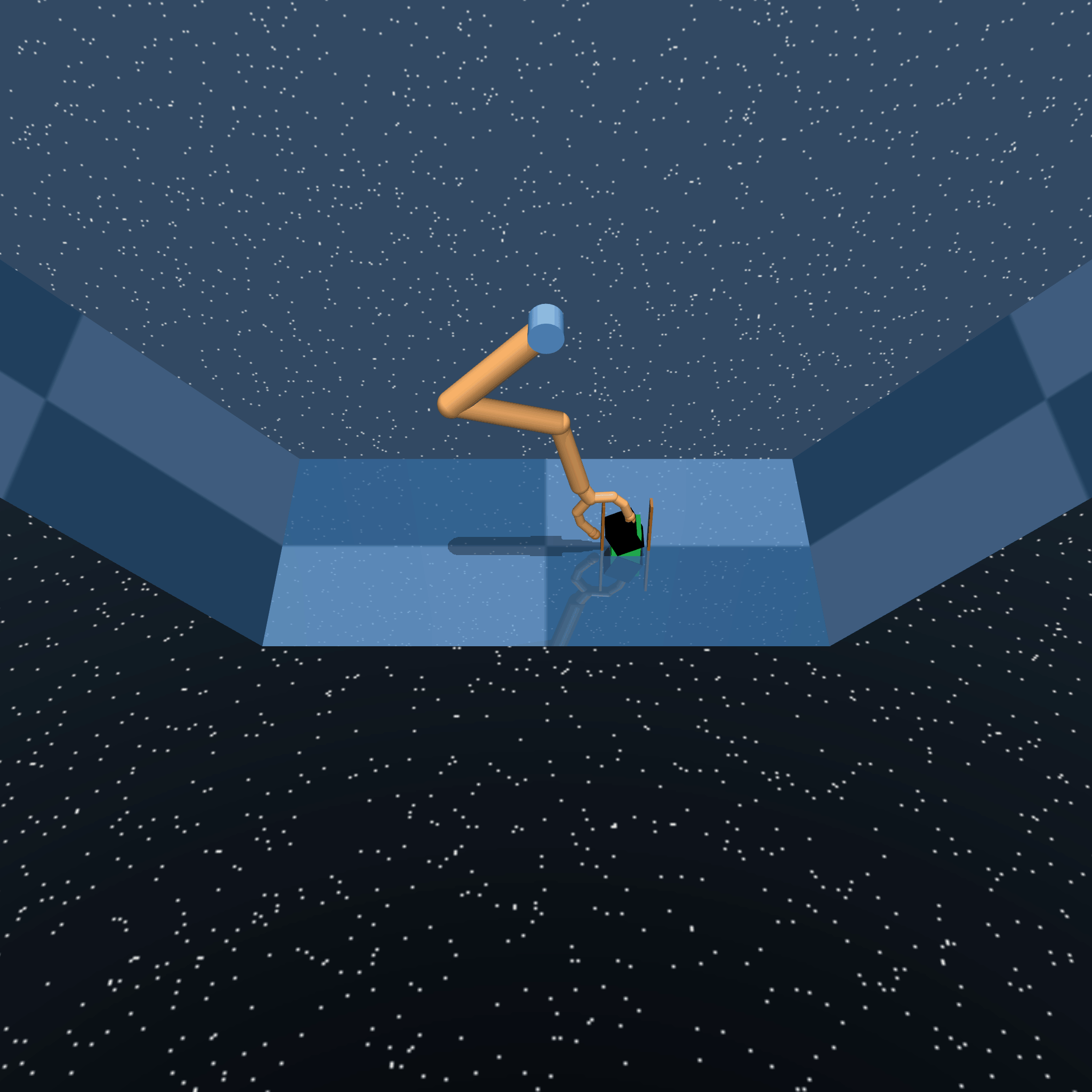}{\caption*{\small Drop Box}}\endminipage\hfill
  \caption{\small \textbf{Pick-Carry-Drop Experiment.} The robot needs to approach, lower to, grasp, pick-up, carry, and drop the box to solve the task.\vspace{-.2cm}}\label{8PP-decompose}
\end{figure*}
\subsection{Using Expert Demonstration as Both the Input Task Context Variables and Training Signal for the Task Encoder}
Learning robust task embeddings enables robots to generalize to new tasks quickly~\cite{rakelly2019efficient}. To this end, our algorithm proposes to use 64 expert timesteps as the input task context variable $c$ into the task encoder, as opposed to 64 robot timesteps. This is because context variables should explore the task and environment sufficiently well to expose the key information of the task, and expert demonstration timesteps are an ideal candidate compared to the timesteps from the robot's suboptimal policy. As a result, the context variable $c$ input into the task encoder only includes the states and actions of the expert, but not the rewards or the next states.

In addition, we choose the BC loss $\mathcal{L}^{BC}$ in Eq.\ref{eq:bc} as the training loss for learning the task encoder $\Psi_\phi$. This BC loss is stable since the expert timesteps are fixed. In contrast, the IRL loss $\mathcal{L}^{IRL}$ (Eq.\ref{eq:discriminatorloss}) and the policy loss $\mathcal{L}^{RL}$ (Eq.\ref{eq:rl}) are less stable because the training data distribution for both losses are non-stationary. This design choice also allows us to learn a robust task embeddings first via BC pre-training before performing meta-IRL training via SQUIRL. We empirically observe that such pre-training can improve the training stability and the sample efficiency of SQUIRL, but the final policy performance is similar with or without BC pre-training. In summary, our algorithm is detailed in Algorithm~\ref{algo:irl} (train) and Algorithm~\ref{algo:irltest} (test), with hyperparameters detailed here\footnote{Hyperparameters in Algorithm~\ref{algo:irl} and~\ref{algo:irltest}. Policy gradient batch size $\mathcal{B}$: 1024 (non-vision), 128 (vision); task embedding batch size $C$: 64; all learning rates: $3e^{-4}$; starting SAC alpha: $1e^{-5}$; SAC target entropy: $-300$; IRL updates per epoch $J$: $400$; policy updates per epoch $K$: $2000$; task embedding size $\mathcal{Z}$: 32; meta-batch size $m$: 10; discount rate $\gamma$: 0.99}.

\section{Experiments and Results Analysis} 
\label{sec:exp}

We evaluate the generality and robustness of our algorithm across long-horizon vision and non-vision tasks with continuous state and action spaces in both simulation (Pick-Carry-Drop, a horizon of 1024 timesteps, 30 train tasks) and real-world (Pick-Pour-Place, a horizon of 100 timesteps, 117 train tasks).
There is only a \textit{single} expert demonstration for \textit{each} of the train or test tasks. 
We compare with the PEARL-BC baseline, which is the behavioral cloning version of  PEARL~\cite{rakelly2019efficient}. \textbf{Evaluation:} We evaluate real-robot and simulation experiments on \textbf{50} and \textbf{500} trials respectively across \textbf{50} seen and unseen tasks. We report mean and standard deviation (``stdev'' hereafter). The performance difference between different experiments is statistically significant if the difference in \textbf{mean} is at least \textbf{either} standard deviation away. Experimental video is at \url{http://crlab.cs.columbia.edu/squirl}. 
\begin{figure*}[h]
\captionsetup{justification=centering}
  \centering
    \minipage{0.162\textwidth}\includegraphics[trim={100 110 240 10}, clip,width=\linewidth,frame]{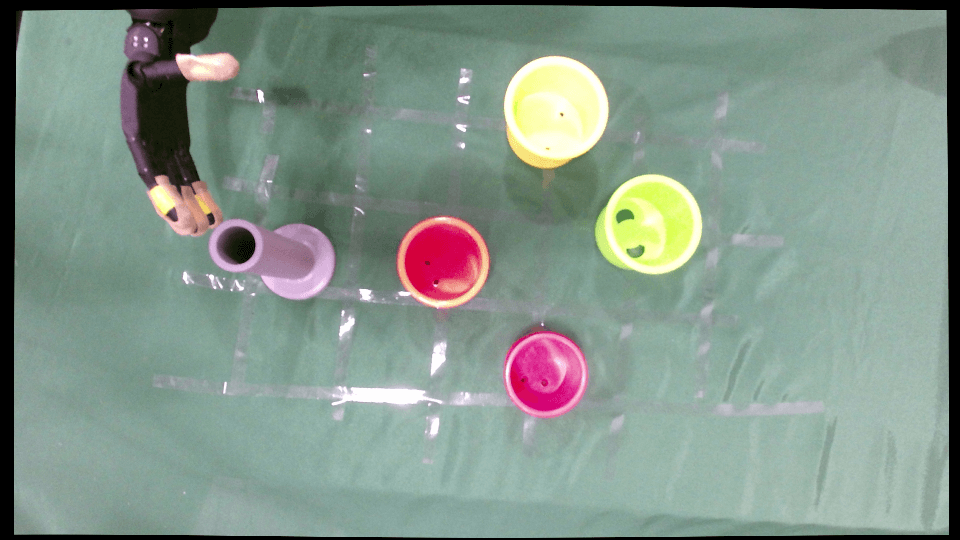}\endminipage\hfill
    \minipage{0.162\textwidth}\includegraphics[trim={100 110 240 10}, clip,width=\linewidth,frame]{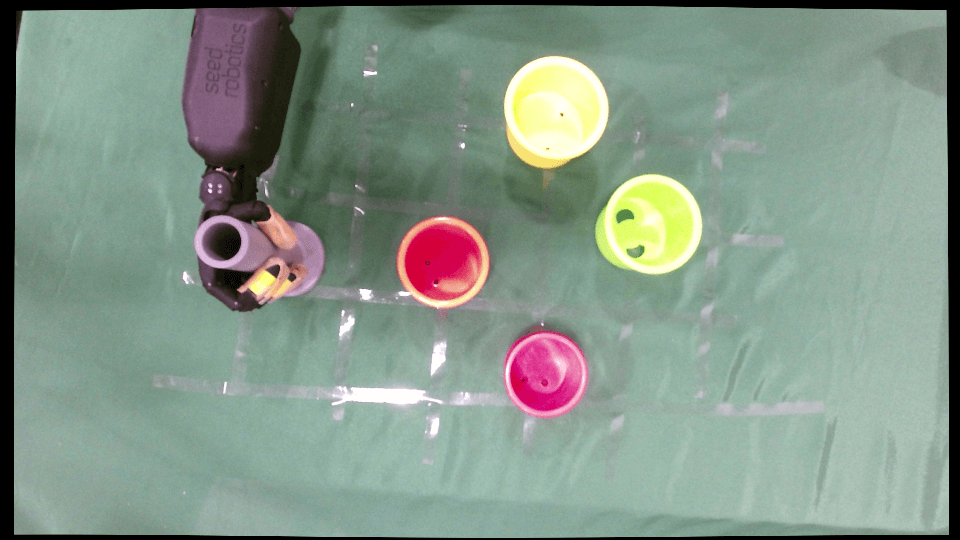}\endminipage\hfill
    \minipage{0.162\textwidth}\includegraphics[trim={100 110 240 10}, clip,width=\linewidth,frame]{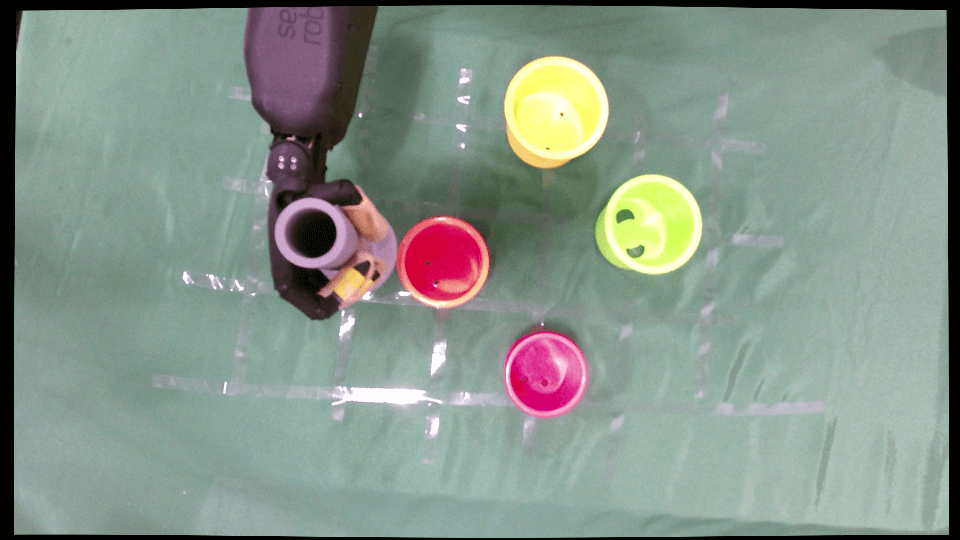}\endminipage\hfill
    \minipage{0.162\textwidth}\includegraphics[trim={100 110 240 10}, clip,width=\linewidth,frame]{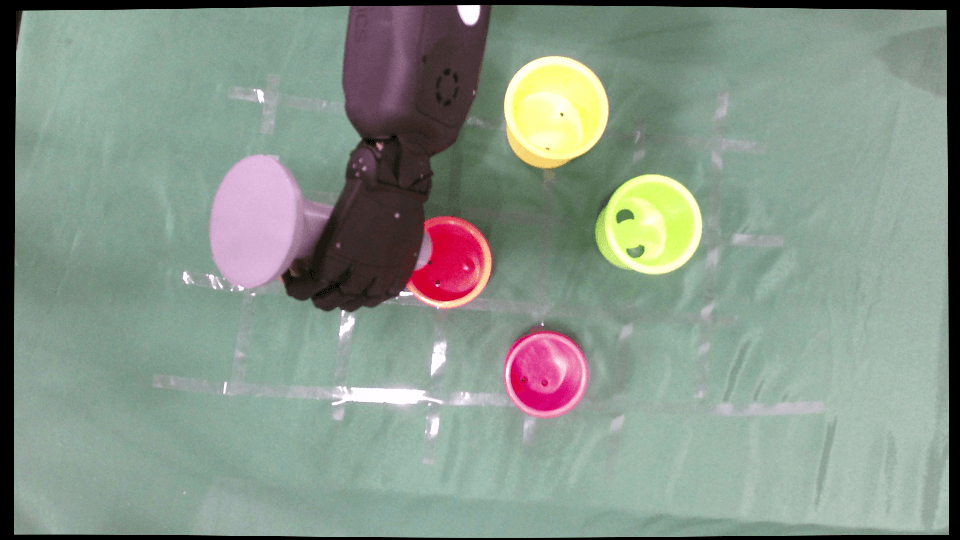}\endminipage\hfill
    \minipage{0.162\textwidth}\includegraphics[trim={100 110 240 10}, clip,width=\linewidth,frame]{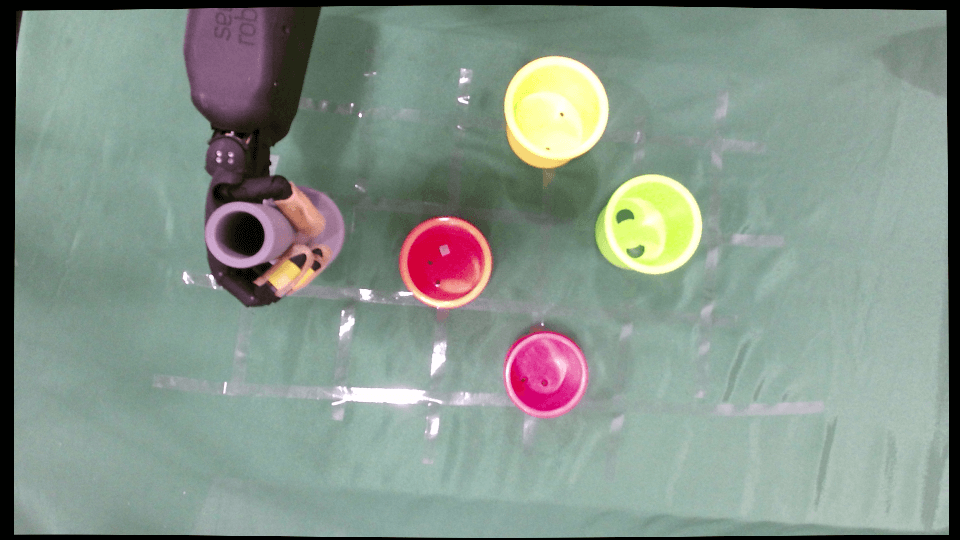}\endminipage\hfill
    \minipage{0.162\textwidth}\includegraphics[trim={100 110 240 10}, clip,width=\linewidth,frame]{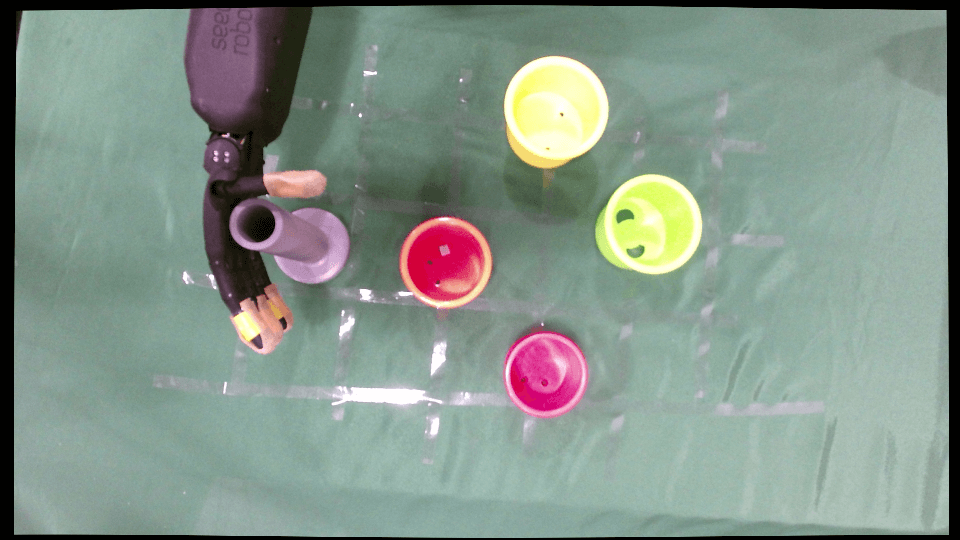}\endminipage\hfill
    \minipage{0.162\textwidth}\includegraphics[trim={50 225 380 0}, clip,width=\linewidth,frame]{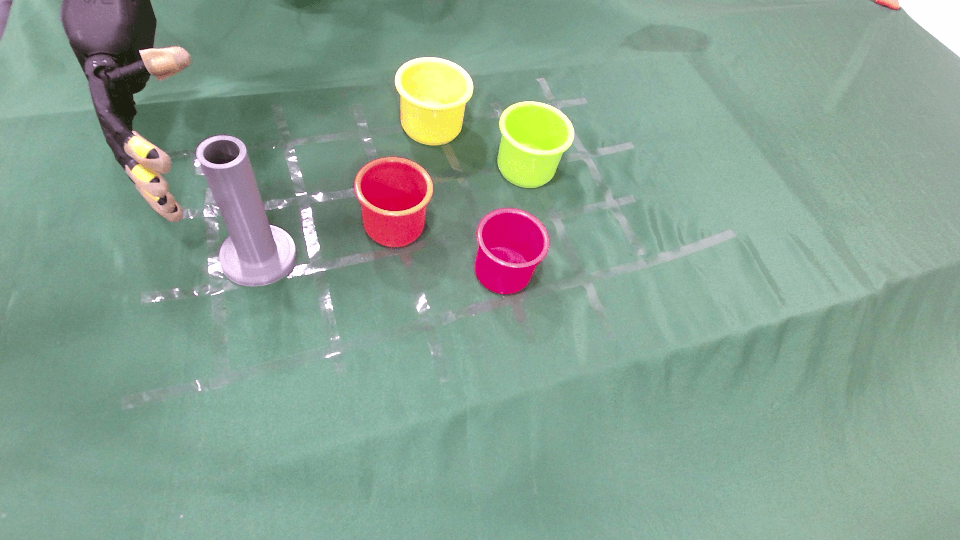}{\caption*{\small Approach Bottle}}\endminipage\hfill
    \minipage{0.162\textwidth}\includegraphics[trim={50 225 380 0}, clip,width=\linewidth,frame]{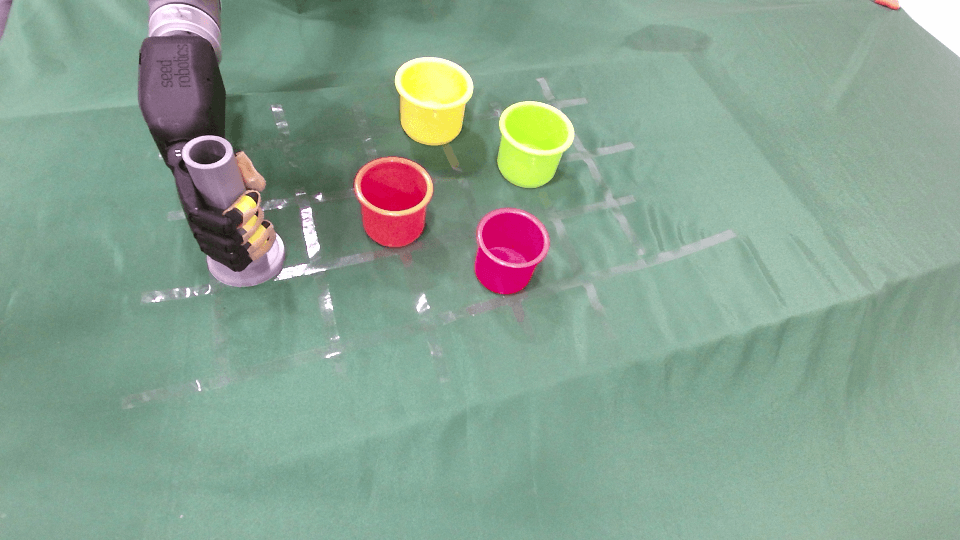}{\caption*{\small Grasp Bottle}}\endminipage\hfill
    \minipage{0.162\textwidth}\includegraphics[trim={50 225 380 0}, clip,width=\linewidth,frame]{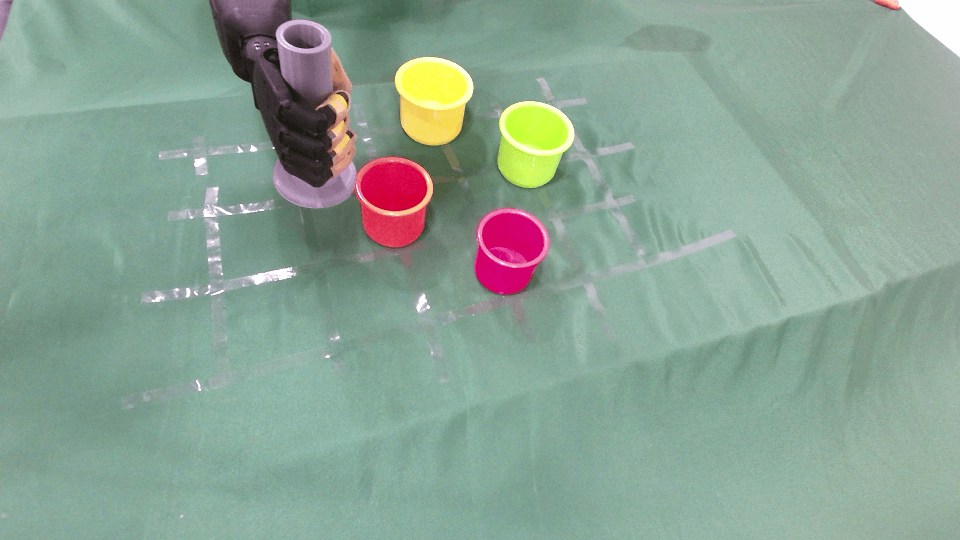}{\caption*{\small Carry Bottle}}\endminipage\hfill
    \minipage{0.162\textwidth}\includegraphics[trim={50 225 380 0}, clip,width=\linewidth,frame]{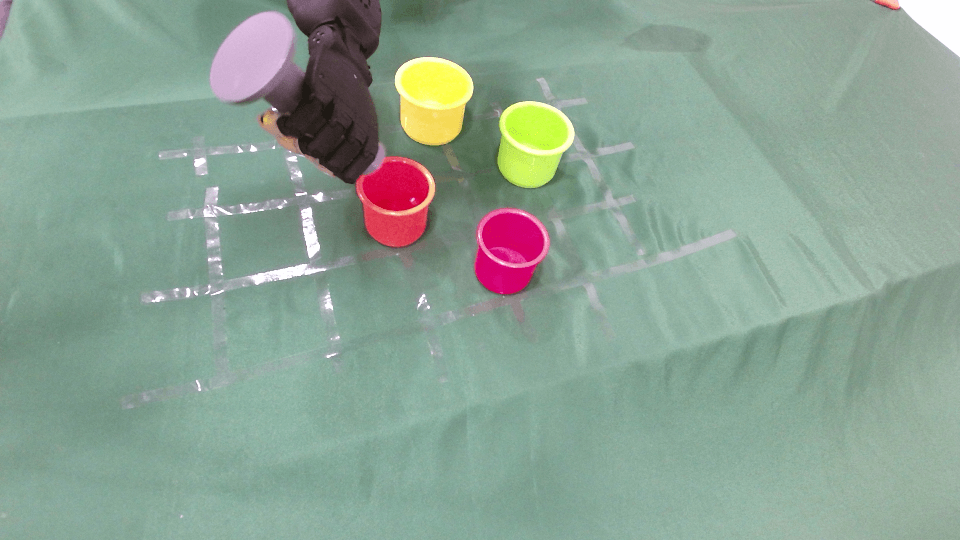}{\caption*{\small Pour \textcolor{orange}{\textbf{Orange}} Cup}}\endminipage\hfill
    \minipage{0.162\textwidth}\includegraphics[trim={50 225 380 0}, clip,width=\linewidth,frame]{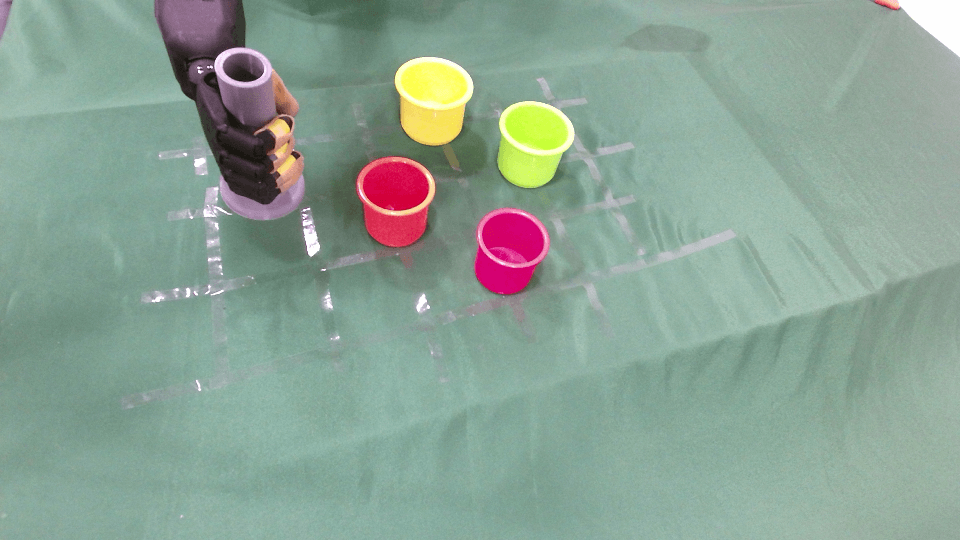}{\caption*{\small Carry Bottle}}\endminipage\hfill
    \minipage{0.162\textwidth}\includegraphics[trim={50 225 380 0}, clip,width=\linewidth,frame]{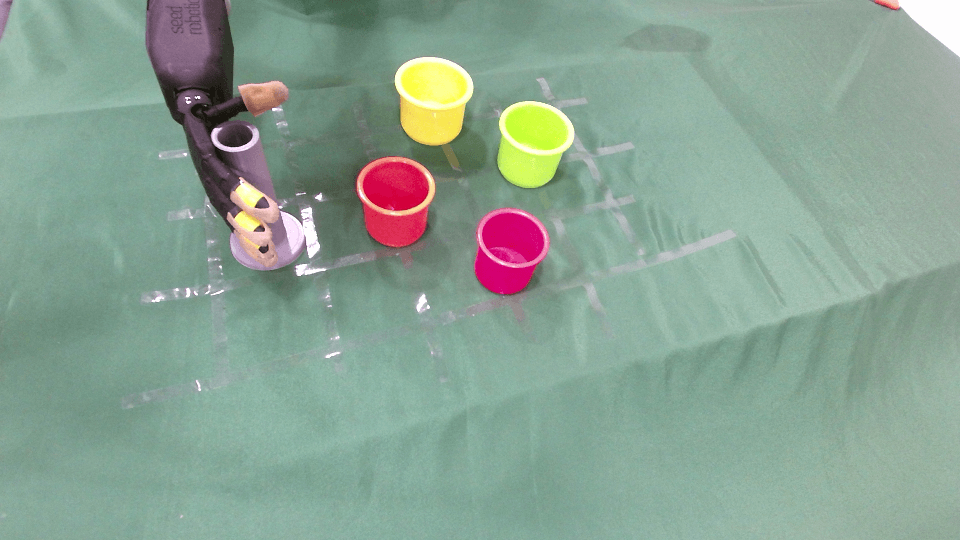}{\caption*{\small Place Bottle}}\endminipage\hfill
  \caption{\small \textbf{Pick-Pour-Place at Test Time.} To solve this task, the robot needs to first approach, grasp and carry the grey bottle, pour the iron pebble inside the bottle into a specific container, and carry and place the bottle back on the table. At the beginning of each task, the bottle is not in hand, and but the iron pebble is already in the bottle. Top row: top-down camera images. Bottom row: 45\degree camera images.\vspace{-.2cm}}\label{ppp}
\end{figure*}
\subsection{Simulation Experiment: Pick-Carry-Drop}
\textbf{Description.} We modify the planar Stacker task~\cite{dmcontrol} to create ``Pick-Carry-Drop''.
Shown in Fig.\ref{8PP-decompose}, a robot is tasked to approach, pick, carry, and drop the black box into the stack marked in green. The task is successful if the box is dropped into the stack within 1024 timesteps, and failed otherwise.

\textbf{State Space.} We evaluate our algorithm on both the vision and the non-vision version of the task, to demonstrate that SQUIRL is \textit{general} across different state space modalities. The state space for the vision version includes 1) the joint angles and velocities for its 5-DOFs, 2) a one-hot vector indicating the current stage of the task, and 3) an RGB image shown in Fig.\ref{8PP-decompose}. The non-vision version's state space replaces the RGB image with the \textit{position of the black box}.

\textbf{Action Space}. The robot controls  its 5-DOF \textit{joint torques}. 

\textbf{Task Definition}. There are a total of 30 training tasks in this experiment, each corresponding to a different \textit{drop location}: $x \in \{-0.15, -0.14, \ldots , 0.14\}$. 
During test time, we randomly sample a new, \textit{real-valued} drop location from the maximum valid range: $x \in [-0.25, 0.25]$. The green drop location is \textit{invisible} in both the vision and the non-vision version of the task. Therefore, the robot needs to infer the green drop location (i.e., task information) solely from the provided expert video demonstration. On the other hand, the starting pose of the robot and the location of the black box are all initialized randomly at the beginning of each task. 

\textbf{Robot Trials}. The robot uses 150 training trials \textit{in total}.

\textbf{Expert Demonstration}. We trained an expert policy from scratch via RL to provide expert demonstrations. The reward function used to train the expert policy comprises of six stages, each stage with a reward of 10. Designing this reward function has taken significant human effort, which exhibits the value of directly learning from video demonstrations. 

\vspace{.2cm}
\begin{table}[h]
\centering
\caption{Pick-Carry-Drop Results (\% Drop Success$\pm$Stdev)}
\label{tab:sim}
\begin{tabular}{|c|c|c|c|c|}
\hline
 Tasks & Seen& Unseen & Seen& Unseen\\ \hline
& \multicolumn{2}{|c|}{Vision} & \multicolumn{2}{|c|}{Non-Vision} \\ \hline
SQUIRL (BC + IRL) & \textbf{95.8$\pm$1.7} & \textbf{95.0$\pm$1.5} & \textbf{97.3$\pm$3.0} & \textbf{96.9$\pm$2.0} \\
Baseline (PEARL-BC) & 77.8$\pm$1.6 & 76.5$\pm$0.7&90.8$\pm$2.5 & 89.5$\pm$1.6 \\\hline
 \multicolumn{5}{|c|}{\textbf{Ablation: }No BC Joint Training or BC Pre-training}\\\hline
SQUIRL (IRL Only)& 93.8$\pm$1.8 & 93.2$\pm$1.6 & 94.7$\pm$1.7 & 93.9$\pm$1.4\\\hline
\end{tabular}
\end{table}

\textbf{Simulation Results and Analysis.} 
As shown in Table~\ref{tab:sim}, our algorithm, ``SQUIRL (BC + IRL)'', pre-trains via BC and then trains the policy using both the BC loss (Eq.\ref{eq:bc}) and the IRL policy gradient loss (Eq.\ref{eq:rl}). It statistically significantly outperforms the PEARL-BC baseline in both the vision (95.8\%$\pm$1.7 vs. 77.8\%$\pm$1.6) and non-vision (97.3\%$\pm$3.0 vs. 90.8\%$\pm$2.5) version of the task for seen tasks. For unseen tasks, we observed similar outperformance (95.0\%$\pm$1.5 vs. 76.5\%$\pm$0.7 in the vision case and 96.9\%$\pm$2.0 vs. 89.5\%$\pm$1.6 in the non-vision case). 
Qualitatively, in the PEARL-BC's case, the robot sometimes misses the drop location as it attempts to drop the box or fails to pick up the box when the box gets stuck by the walls of the stack (kindly see website). 
The performance drop of the baseline from the non-vision version (90.8\%$\pm$2.5 and 89.5\%$\pm$1.6 for seen and unseen tasks) to the vision version (77.8\%$\pm$1.6 and 76.5\%$\pm$0.7 for seen and unseen tasks) is mainly because vision-based manipulation tends to suffer from larger compounding errors. 
Nevertheless, as evident in the statistical similarities between seen and unseen tasks for SQUIRL (95.8\%$\pm$1.7 vs. 95.0\%$\pm$1.5 for vision) and PEARL-BC (77.8\%$\pm$1.6 vs. 76.5\%$\pm$0.7 for vision), both algorithms can generalize to unseen tasks, due to the generalizability of task embeddings.

\textbf{Ablation: IRL Gradient Only}. To compare the performance contribution of SQUIRL's meta-IRL core training procedure directly against PEARL-BC,
we created ``SQUIRL (IRL only)'', which trains the policy using only the policy gradient loss in Eq.\ref{eq:rl} (no BC joint training or pre-training). This \textit{ablated} version still outperforms the PEARL-BC baseline (93.8\%$\pm$1.8 vs. 77.8\%$\pm$1.6 for seen vision tasks, 93.2\%$\pm$1.6 vs. 76.5\%$\pm$0.7 for unseen vision tasks). Nevertheless, by combining BC and IRL gradients, ``SQUIRL (BC + IRL)'' improves performance slightly further (95.8\%$\pm$1.7 and 95.0\%$\pm$1.5). Intuitively, while BC only matches the expert's conditional action distribution under the \textit{expert's } state distribution, BC's supervised learning signal is stabler than IRL. Joint training with BC and IRL gradients can be interpreted as combining the stability of BC and the robustness of Q-functioned IRL, by matching the conditional action distribution of the expert under the broader state distribution of the expert-robot mixture experience (Eq.\ref{eq:indirect}), in addition to matching the expert's joint state-action distribution (Eq.\ref{eq:joint}).

\subsection{Real-Robot Experiment: Pick-Pour-Place}
\textbf{Description.} We evaluated our algorithm on the UR5-Seed robot (Fig.\ref{fig:hardware}) to perform a set of long-horizon pick-pour-place tasks. As shown in Fig.\ref{fig:hardware}, in each task, there is a grey cylindrical bottle, an iron pebble that is already in the bottle, and more than one container on the table. The robot is tasked to approach and pick-up the grey bottle, pour the iron pebble into a specific container, and place the bottle back on the table. The task is a success only if the pebble is poured into the \textit{correct} container and the bottle is placed upright on the table within $H=100$ timesteps, and a failure otherwise. 

\textbf{State Space.} The state space contains a top-down or 45\degree camera's RGB image (Fig.\ref{ppp}), and 2 binary indicators for whether the robot has poured or closed the hand, respectively.

\textbf{Action Space.} The action space includes the Cartesian unit directional vector for the end-effector movement. During each timestep, the robot can adjust the end-effector by 2cm along any 3D direction. The action space also includes a binary indicator to control the arm vs. the hand and a trinary indicator to close, open, or rotate the hand for pouring. 

\textbf{Orthogonality to State and Action Representions}. While Pick-Pour-Place can be tackled by first localizing the correct container via object detection (alternative state space) and then executing motion-planning trajectories to pour (alternative action space), our algorithm is \textit{general} across and orthogonal to alternative state and action spaces.

\textbf{Task Definition.} As shown in each row of images in Fig.\ref{fig:intro}, each task is defined by the positions and colors of the containers, and by the correct container to pour into. There are \textit{always} \textit{only} the green and yellow containers in the 117 train tasks. 25 of the 50 test tasks have the green and yellow containers at \textit{new} positions. The remaining 25 test tasks \textit{add} the red and the orange \textit{unseen} containers, or either. Since there is always more than one container in the RGB image, the robot will not know which container to pour into \textit{without} the expert demonstration. Therefore, the robot needs to depend solely on the task encoder's ability to extract the correct task information from the expert demonstration.

\textbf{Robot Trials}. The robot collects 90 training trials \textit{in total}.

\textbf{Expert Demonstration.} We collect demonstrations via teleoperation using a Flock of Birds sensor\footnote{Flock of Birds is a 6D pose tracker from Ascension Technologies Corp.}. Using the human wrist pose detected by the sensor in real-time, we move, open, close, or rotate the robot hand for pouring. We collected $117$ video demonstrations across 117 tasks for training. It takes 1-2 minutes to collect one demonstration.

\vspace{.2cm}
\begin{table}[h]
\centering
\caption{Pick-Pour-Place Results (\% Pour Success$\pm$Stdev)}
\label{tab:real}
\begin{tabular}{|c|c|c|c|}
\hline
 Tasks & RGB Image & Seen& Unseen\\ \hline
SQUIRL (BC + IRL) & \multirow{3}{*}{Top-Down (90\degree)} &\textbf{92.0$\pm$4.5} & \textbf{90.0$\pm$7.1} \\
Baseline (PEARL-BC) & &70.0$\pm$7.1 & 68.0$\pm$11.0\\
Baseline (Standard-BC)& & 60.0$\pm$10.0 &  56.0$\pm$11.4\\\hline
SQUIRL (BC + IRL) &$45\degree$ (\textbf{Ablation})& 90.0$\pm$7.1 & 88.0$\pm$8.4 \\ \hline
\end{tabular}
\end{table}

\textbf{Real-robot Results and Analysis.} As shown in Table~\ref{tab:real}, our algorithm outperforms the PEARL-BC baseline statistically significantly in both seen tasks (92.0\%$\pm$4.5 vs. 70.0\%$\pm$7.1) and unseen tasks (90.0\%$\pm$7.1 vs. 68.0\%$\pm$11.0). This observed outperformance mainly originates from our soft Q-functioned IRL formulation, which forces the robot to imitate the expert under a much wider state distribution provided by the expert-robot mixture trajectories, instead of the narrow state distribution of the expert demonstrations. This helps reduce compounding errors during task execution. The low performance of the PEARL-BC baseline is mainly due to additional compounding errors induced by real-world sensory noises such as unstable lighting conditions and small perturbation to camera positions. Qualitatively, the PEARL-BC baseline sometimes pours into the wrong container, misses the target container by a few centimeters, or moves past the target container while failing to pour in time (kindly see website for examples). Nevertheless, from the statistical similarity between seen and unseen tasks for both our algorithm (92.0\%$\pm$4.5 vs. 90.0\%$\pm$7.1) and PEARL-BC (70.0\%$\pm$7.1 vs. 68.0\%$\pm$11.0), we see that the learned task encoder is still effectively generalizing to a new, related task.

\textbf{Comparison to the ``Standard-BC'' Baseline}. We also compared to ``Standard-BC'' (60.0\%$\pm$10.0 and 56.0\%$\pm$11.4 for seen and unseen tasks), which performs no meta-learning and learns every train or test task \textit{independently} from scratch via BC. As a result, the neural network \textit{overfits} to the single demonstration and fails to generalize to real-world sensory (camera) noises at test time. Note that Standard-BC's unseen-task performance is slightly lower than seen tasks since the unseen tasks are more challenging with at most 4 containers on the table, compared to only 2 containers in seen tasks.

\textbf{Ablation: Non-top-down Camera}. We also tested our algorithm with a $45\degree$  RGB image (90.0\%$\pm$7.1 and 88.0\%$\pm$8.4 for seen and unseen tasks) against a top-down RGB image (92.0\%$\pm$4.5 and 90.0\%$\pm$7.1 for seen and unseen tasks). The statistical similarity between the two shows that SQUIRL is \textit{general} and can accept a non-top-down RGB input image.

\section{Conclusion}
We introduced SQUIRL, a robust, efficient, and general Soft Q-functioned meta-IRL algorithm, towards enabling robots to learn from limited expert (one per task) and robot (90 in total) trajectories. This algorithm is statistically significantly more robust than behavioral cloning and requires no trial-and-errors at test time. Finally, this general algorithm has been tested to work with various long-horizon manipulation tasks, and across vision and non-vision state and action spaces. In the future, we will extend this algorithm to learn from direct human-arm demonstrations instead of teleoperation. This will lower the cost of collecting real-world expert demonstrations further. We also aim to incorporate hierarchical learning into SQUIRL to solve much longer horizon manipulation tasks by reusing low-level subpolicies.

\bibliographystyle{IEEEtran}
\footnotesize\bibliography{IEEEabrv}
\end{document}